\PassOptionsToPackage{unicode}{hyperref}
\PassOptionsToPackage{hyphens}{url}
\PassOptionsToPackage{dvipsnames,svgnames,x11names}{xcolor}
\documentclass[
  11pt,
]{article}
\usepackage{xcolor}
\usepackage[margin=2.5cm]{geometry}
\usepackage{amsmath,amssymb}
\usepackage{iftex}
\ifPDFTeX
  \usepackage[T1]{fontenc}
  \usepackage[utf8]{inputenc}
  \usepackage{textcomp} % provide euro and other symbols
\else % if luatex or xetex
  \usepackage{unicode-math} % this also loads fontspec
  \defaultfontfeatures{Scale=MatchLowercase}
  \defaultfontfeatures[\rmfamily]{Ligatures=TeX,Scale=1}
\fi
\usepackage{lmodern}
\ifPDFTeX\else
\fi
\IfFileExists{upquote.sty}{\usepackage{upquote}}{}
\IfFileExists{microtype.sty}{% use microtype if available
  \usepackage[]{microtype}
  \UseMicrotypeSet[protrusion]{basicmath} % disable protrusion for tt fonts
}{}
\makeatletter
\@ifundefined{KOMAClassName}{% if non-KOMA class
  \IfFileExists{parskip.sty}{%
    \usepackage{parskip}
  }{% else
    \setlength{\parindent}{0pt}
    \setlength{\parskip}{6pt plus 2pt minus 1pt}}
}{% if KOMA class
  \KOMAoptions{parskip=half}}
\makeatother
\usepackage{longtable,booktabs,array}
\usepackage{caption}
\usepackage{calc} % for calculating minipage widths
\usepackage{etoolbox}
\makeatletter
\patchcmd\longtable{\par}{\if@noskipsec\mbox{}\fi\par}{}{}
\makeatother
\usepackage{graphicx}
\makeatletter
\newsavebox\pandoc@box
\newcommand*\pandocbounded[1]{% scales image to fit in text height/width
  \sbox\pandoc@box{#1}%
  \Gscale@div\@tempa{\textheight}{\dimexpr\ht\pandoc@box+\dp\pandoc@box\relax}%
  \Gscale@div\@tempb{\linewidth}{\wd\pandoc@box}%
  \ifdim\@tempb\p@<\@tempa\p@\let\@tempa\@tempb\fi% select the smaller of both
  \ifdim\@tempa\p@<\p@\scalebox{\@tempa}{\usebox\pandoc@box}%
  \else\usebox{\pandoc@box}%
  \fi%
}
\def\fps@figure{htbp}
\makeatother
\providecommand{\tightlist}{%
  \setlength{\itemsep}{0pt}\setlength{\parskip}{0pt}}
\usepackage[round]{natbib}
\usepackage{etoolbox}
\usepackage{needspace}
\AtBeginEnvironment{longtable}{\footnotesize\setlength{\tabcolsep}{3pt}}
\usepackage{bookmark}
\IfFileExists{xurl.sty}{\usepackage{xurl}}{} % add URL line breaks if available
\makeatletter
\@ifundefined{xmpquote}{}{}
\makeatother
\hypersetup{
  pdftitle={LightOnOCR for AMNLT: Reading Chant Notation and Lyrics in One Pass{,} from Single Staves to Whole Pages},
  colorlinks=true,
  linkcolor={Maroon},
  filecolor={Maroon},
  citecolor={Blue},
  urlcolor={Blue},
  pdfcreator={LaTeX via pandoc}}

\title{LightOnOCR for AMNLT: Reading Chant Notation and Lyrics in One
Pass, from Single Staves to Whole Pages}
\author{Robert Klugseder\\ {\small Austrian Centre for Digital Humanities (ACDH), Austrian Academy of Sciences, Vienna}}
\date{}

\begin{document}
\maketitle
\begin{abstract}
Aligned music notation and lyrics transcription (AMNLT) asks for the
notes of a chant, its sung text and the assignment of syllables to notes
in one output. Fuentes-Martínez et al., who defined the task, found that
the one vision-language model they tested fell far behind specialised
recognisers. We fine-tune LightOnOCR-3-1B, a page OCR model with one
billion parameters, end to end, vision encoder included, to write a
staff as a GABC string whose letters denote staff positions. Trained on
the published AMNLT training split alone, it more than halves the best
published aligned music and lyrics error rate (AMLER) on Solesmes, the
one directly comparable dataset (9.9 against 21.0 \%); on Einsiedeln and
Salzinnes, measured in a different encoding, its figures are also lower.
The benchmark does not measure generalisation to unseen manuscripts:
while the pitch error on AMNLT validation stays between 1.7 and 1.9 \%,
on 200 staves from five unseen manuscripts it falls from 24.9 to 1.3 \%
as we add chant aligned through CantusDB, Transkribus ground truth and
silver staves kept only where two readers agree. Trained also on
multi-staff images, it reads whole pages without staff detection (pitch
error 3.3 to 3.8 \% on validation and test pages, 1.9 \% on pages of
unseen manuscripts). Remaining errors include Gothic F clefs read as C
clefs, notes dropped with abbreviated or illegible text and lower
accuracy on pages with small notation. Controlled ablations are needed
to explain the gains.
\end{abstract}

\section{Introduction}\label{introduction}

A chant book carries two things at once: a melody written in neumes on a
staff and the text that is sung to it. For editing, searching or
performing the chant, neither is of much use without the other, and both
are of little use without knowing which notes belong to which syllable.
Fuentes-Martínez et al.~gave this problem a name and a formulation,
aligned music notation and lyrics transcription (AMNLT), together with
four datasets, a family of methods and two metrics that separate what
was recognised from how it was aligned \citep{fuentes2026amnlt}. Their
study is the foundation of the present one. It reaches three conclusions
that we take as our starting point. Recognising music and text
separately and aligning them afterwards gives the best transcription and
loses in the alignment. End-to-end models that write the aligned
sequence directly align better, and among them the image-to-sequence
models (Sheet Music Transformer, DAN) are the strongest where there is
enough training data. The one vision-language model in the study,
PaliGemma 2 adapted with LoRA, stays at an AMLER between 47 and 70 \% on
all four datasets, and the authors conclude that ``as of today,
vision-language models seem unable to achieve competitive results for
the AMNLT task''.

This article asks whether that last conclusion holds for a different
vision-language model trained in a different way. LightOnOCR-3-1B is a
page OCR model with about one billion parameters: a vision transformer,
a projector and a small language-model decoder, pre-trained to turn
document pages into text \citep{lightonocr2026, lightonocr3card}. We
fine-tune all of it, the vision encoder included, on staves of chant
with their text line, and let it write one string per staff in the
GABC-like encoding that AMNLT uses for its music-aware targets.

The choice of model was prompted by a variant of LightOnOCR-3-1B
fine-tuned on German Kurrent script of the 19th century
\citep{straub2026historichtr}\footnote{\url{https://github.com/b-straub/historic-htr-ocr}},
which the author had extended by further fine-tuning to late-medieval
bastarda and early German Kurrent with surprisingly good results.

On the benchmark, the answer comes from the first run, trained on
nothing but the AMNLT training split. On Solesmes, the one directly
comparable dataset, it more than halves the best published AMLER (9.92
against 20.98 \%); on Einsiedeln and Salzinnes its figures are also
lower, but they are measured in a different encoding (Section 7.1). The
larger part of the work and of this article follows from there. The
benchmark turned out to say little about manuscripts the model has not
seen: the same first run reads five held-out manuscripts in Gothic and
square notation with a pitch error of 24.9 \%. The architecture stayed
as it was; as that figure came down to 1.3 \%, what changed from run to
run was the training data: chant aligned with its text through the
Cantus Database, ground truth from Transkribus, and staves whose notes
were read by two different recognisers and kept only where both agree.
All of it could be pooled because all of it is written in one encoding.
Cropping single staves turned out to be a source of error in its own
right: where staves are tightly spaced, a crop easily includes notes of
the neighbouring staff, which the model then transcribes as well
(Section 8.4), and the image filters that exclude such crops discard 62
\% of the candidates of the chant corpus (Section 4.2). The second step
was therefore to stop cutting pages into staves. A model that reads the
whole page needs neither a staff detector nor cropping at inference.

\begin{figure}
\centering
\pandocbounded{\includegraphics[keepaspectratio,alt={Overview. Sources
of different kinds are brought into one encoding; the training lists
hold single staves and images with several staves; all weights of the
model are trained; the reading is one line per staff. Readings of the
model and of a second reader, where they agree, return as silver staves
(dashed).}]{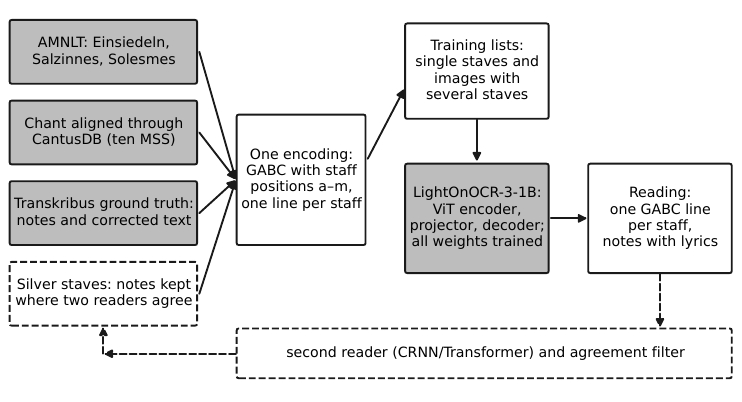}}
\caption{Overview. Sources of different kinds are brought into one
encoding; the training lists hold single staves and images with several
staves; all weights of the model are trained; the reading is one line
per staff. Readings of the model and of a second reader, where they
agree, return as silver staves (dashed).}
\end{figure}

\needspace{10\baselineskip}

The contributions are the following.

\begin{enumerate}
\def\labelenumi{\arabic{enumi}.}
\tightlist
\item
  Joint recognition of chant notation and lyrics with a fully fine-tuned
  page OCR model. With the vision encoder trained and a strip layout
  that keeps a staff step (half the distance between two adjacent staff
  lines) above the patch size, LightOnOCR-3-1B more than halves the best
  published AMLER on Solesmes, the one directly comparable AMNLT test
  set, with the AMNLT training data alone (Section 7.1).
\item
  Cross-manuscript evaluation and a shared data pipeline. The AMNLT
  benchmark saturates, while the pitch error on 200 staves from five
  unseen manuscripts falls from 24.9 to 1.3 \% as data in the target
  notations are pooled in one position-based GABC encoding: chant
  aligned through CantusDB, Transkribus ground truth and silver staves
  from a two-reader filter that also exposed errors in the human ground
  truth. The encoding and the second reader are prior work on which we
  build (Sections 3, 4 and 7.2).
\item
  Whole-page recognition without a staff detector at inference, with an
  analysis of resolution and of the failure modes, among them notes that
  fall together with abbreviated or illegible text (Sections 7.3 and 8).
\end{enumerate}

\section{Related Work}\label{related-work}

\textbf{Aligned transcription of music and lyrics.} Our account of the
work before AMNLT, and of the Sheet Music Transformer below, follows the
AMNLT study \citep[Sections 2 and 7.4]{fuentes2026amnlt}; we have not
checked it against the works themselves. According to that account, the
interplay of music and text in vocal sources was used before AMNLT
mainly to improve separate music and text recognition
\citep{villarreal2024enhancing}, and a first holistic model for aligned
output was trained on synthetic data only
\citep{martinezsevilla2023holistic}. \citet{fuentes2026amnlt} compare a
divide-and-conquer baseline (two convolutional recurrent networks
trained with connectionist temporal classification, CRNN-CTC, with
post-alignment) with three end-to-end families: a base holistic
CRNN-CTC, unfolding methods, and language modelling, the last comprising
the Sheet Music Transformer \citep{riosvila2024smt}, DAN
\citep{coquenet2023dan}, TrOCR \citep{li2023trocr} and PaliGemma 2
\citep{steiner2024paligemma2} with LoRA \citep{hu2022lora}. Their
metrics AMLER and AlER build on the bag-of-words error rate of
\citet{vidal2023bwer}. A recent article treats recognition and alignment
of historical musical documents at the level of the full page
\citep{villarreal2026fullpage}.

\textbf{End-to-end optical music recognition.} The Sheet Music
Transformer transcribes complex scores as an image-to-sequence task
\citep{riosvila2024smt}; its successor reads full pages of pianoform
music without prior layout analysis, after curriculum pre-training on
synthetic data \citep{riosvila2024smtpp}. MuSViT is a vision transformer
pre-trained with masked autoencoding on 9.7 million pages of sheet music
and evaluated, among other tasks, on full-page and staff-level
recognition \citep{penarrubia2026musvit}. Both lines of work point in
the direction taken here, pre-training and page-level reading, from the
side of music; we come from the side of document OCR.

\textbf{Optical recognition of medieval notation.} For square notation
and chant there is a body of work on staff, symbol and melody detection
and on complete pipelines \citep{wick2019staff, hartelt2024pipeline},
and a recent proposal for a foundational model for diastematic Gregorian
notation \citep{kurek2026foundational}.

\textbf{Vision-language models for OCR.} LightOnOCR is an end-to-end
vision-language model for document OCR \citep{lightonocr2026}; the 1B
model used here combines a Pixtral vision encoder
\citep{agrawal2024pixtral} with a Qwen3 decoder \citep{yang2025qwen3}.
The AMNLT paper evaluated PaliGemma 2 in its 3B variant at 896 px, with
LoRA of rank 8 in the language model and the vision block and projector
frozen \citep[Sections 7.4.2 and 7.5]{fuentes2026amnlt}.

\textbf{Encodings and data.} gabc is the text format of the Gregorio
project \citep{gregorio}. For related work on a systematic GABC encoding
of Gregorian chant, see \citet{thomae2024gabc}. The GabcParser of Daniel
Kurek converts between the GABC variants of the AMNLT datasets and
defines the common encoding we use \citep{kurek_gabcparser}. Melodies
and texts of the chant corpus come from the Cantus Database
\citep{cantusdb}, where melodies are stored in Volpiano
\citep{helsen2011volpiano}. The step from those Volpiano transcriptions
to image-level ground truth for ten manuscripts, about 30,000 staff
lines with targets in staff positions, is described in a separate
article \citep{klugseder2026volpiano}; the present article starts from
that ground truth and adds the lyrics and their alignment. Further
ground truth was produced in Transkribus \citep{kahle2017transkribus}
and exported as PAGE XML \citep{pletschacher2010page}.

\textbf{Joint work.} This article forms part of a collaboration of the
PRAIG group and the Austrian Academy of Sciences on late-medieval
plainchant sources, which has also produced a recognition tool with
separate models for layout, music and lyrics
\citep{calvozaragoza2026browser}.

\section{Task and Encoding}\label{task-and-encoding}

\textbf{Output.} For an image of one staff with its text line, the model
writes one string in which syllables and their notes alternate: each
syllable is followed by a pair of parentheses that holds the notes sung
to it. A clef, a bar or a group of notes without text stands in
parentheses of its own. For an image with several staves the model
writes one such line per staff, from top to bottom. Following the
music-aware encoding of AMNLT \citep[Section 5.2]{fuentes2026amnlt},
every character inside the parentheses carries the prefix
\texttt{\textless{}m\textgreater{}}, because gabc uses the same letters
for notes and for text. Figure 2 shows a staff with its target.

\begin{figure}
\centering
\pandocbounded{\includegraphics[keepaspectratio,alt={A staff from AMNLT
Einsiedeln (training split, item 1) with its target in the common
encoding, the same target as stored with the prefix \textless
m\textgreater, and the pseudo-gabc target of the AMNLT paper. In the
common encoding the letters name lines and spaces of the staff and are
read together with the clef; in pseudo-gabc each note is a pitch name
with octave. Rubrics and text without notes are in the image but not in
the target. Image: PRAIG/Einsiedeln\_staffLevel, CC BY-NC
4.0.}]{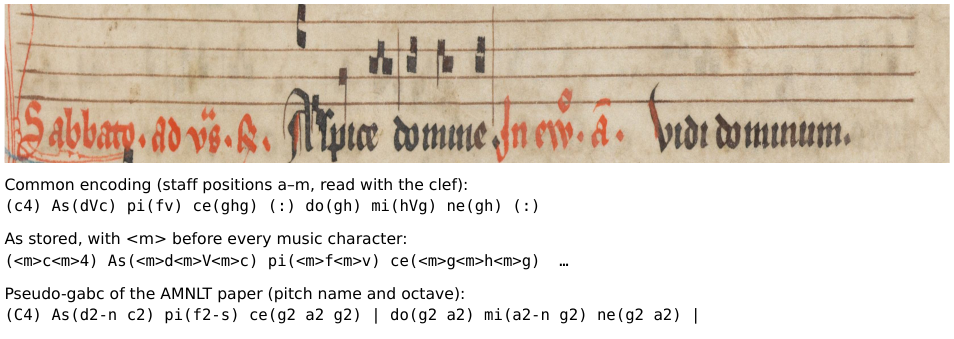}}
\caption{A staff from AMNLT Einsiedeln (training split, item 1) with its
target in the common encoding, the same target as stored with the prefix
\protect\texttt{\textless{}m\textgreater{}}, and the pseudo-gabc target
of the AMNLT paper. In the common encoding the letters name lines and
spaces of the staff and are read together with the clef; in pseudo-gabc
each note is a pitch name with octave. Rubrics and text without notes
are in the image but not in the target. Image:
PRAIG/Einsiedeln\_staffLevel, CC BY-NC 4.0.}
\end{figure}

\textbf{Staff positions instead of pitch names.} The AMNLT paper uses
two encodings. GregoSynth and Solesmes are written in music-aware gabc.
Einsiedeln and Salzinnes are written in a pseudo-gabc derived from MEI,
in which a note is a pitch name with its octave and optional suffixes,
for example \texttt{(C4)\ de(g2\ a2)\ us(g2-n\ f2)}. We train and
measure all datasets in one encoding, the common GABC encoding defined
in the GabcParser \citep{kurek_gabcparser}, in which Einsiedeln and
Salzinnes are converted to the position letters of gabc: the example
becomes \texttt{(c4)\ de(gh)\ us(gVf)}. The letters \texttt{a} to
\texttt{m} name the lines and spaces of the staff from bottom to top
(\texttt{d}, \texttt{f}, \texttt{h}, \texttt{j} are the four lines) and
are read together with the clef. They are geometric: the same pitch has
a different letter under a different clef. This encoding is not a
contribution of this article. It was chosen before the experiments
began, and its value for us is that every other source could be brought
into it: the chant corpus, whose ground truth records staff positions
\citep{klugseder2026volpiano}, the Transkribus ground truth, which also
records positions, and the readings of the second recogniser used in
Section 4.4.

\textbf{Consequences.} The position encoding separates clef and notes.
The model writes what it sees, and a misread clef costs one token, not
every note of the line. That is an advantage in learning and a caveat in
measuring: MER and AMLER on this encoding barely punish a wrong clef,
although the line is then musically displaced, whereas in pseudo-gabc a
wrong clef changes every pitch name. We therefore report the initial
clef as a separate figure throughout. Whether the encoding also raises
accuracy is not measured (Section 9, H4).

\textbf{What the encoding does not unify.} Pitches are encoded uniformly
across our sources; shape signs are not. The sign \texttt{v} (a virga,
in the original encodings a note with the suffix \texttt{-s}) occurs
only in the AMNLT targets, on 37.3 \% of the notes of Einsiedeln, 16.6
\% of Salzinnes and 20.5 \% of Solesmes; none of our own corpora carries
a single one, because the note code of our ground truth does not
distinguish the shape. The sign \texttt{!} for a neume boundary inside a
syllable occurs only in Solesmes. MER across corpus boundaries must be
read with this in mind; our main measure is therefore the pitch error
defined in Section 6.

\section{Data}\label{data}

Table 1 lists the training blocks of the default model and the sets used
for selection and for testing. All targets are in the common encoding
and are checked against its grammar line by line where they were built
by us. The author performed the visual checks and corrections reported
below.

Model names follow one scheme (Table 3). S marks models trained on
single staves, P models also trained on images with several staves. In
the S names, C3 and C10 stand for chant from three and ten manuscripts
of the CantusDB corpus, w for AMNLT targets with word boundaries and T
for the Transkribus corpus. P-ft continues the last staff model; P-b to
P-e are trained from the published weights. P-b produced the readings
for the silver staves (Section 4.4), P-d is the default model reported
throughout, and P-e is a later exploratory run that was not adopted
(Section 8.2).

\needspace{14\baselineskip}

\begin{longtable}[]{@{}
  >{\raggedright\arraybackslash}p{(\linewidth - 6\tabcolsep) * \real{0.2686}}
  >{\raggedright\arraybackslash}p{(\linewidth - 6\tabcolsep) * \real{0.3543}}
  >{\raggedleft\arraybackslash}p{(\linewidth - 6\tabcolsep) * \real{0.2457}}
  >{\raggedright\arraybackslash}p{(\linewidth - 6\tabcolsep) * \real{0.1314}}@{}}
\caption{Training data of the default model P-d (examples per epoch) and
evaluation sets. ``Gold'' means notes transcribed by people and text
corrected by people; ``silver'' means notes read by two recognisers that
agree, with text corrected by people.}\tabularnewline
\toprule\noalign{}
\begin{minipage}[b]{\linewidth}\raggedright
Block or set
\end{minipage} & \begin{minipage}[b]{\linewidth}\raggedright
Origin
\end{minipage} & \begin{minipage}[b]{\linewidth}\raggedleft
Size
\end{minipage} & \begin{minipage}[b]{\linewidth}\raggedright
Role
\end{minipage} \\
\midrule\noalign{}
\endfirsthead
\toprule\noalign{}
\begin{minipage}[b]{\linewidth}\raggedright
Block or set
\end{minipage} & \begin{minipage}[b]{\linewidth}\raggedright
Origin
\end{minipage} & \begin{minipage}[b]{\linewidth}\raggedleft
Size
\end{minipage} & \begin{minipage}[b]{\linewidth}\raggedright
Role
\end{minipage} \\
\midrule\noalign{}
\endhead
\bottomrule\noalign{}
\endlastfoot
AMNLT, word boundaries added & PRAIG staff-level datasets Einsiedeln,
Salzinnes, Solesmes \citep{fuentes2026amnlt} & 4,633 (3,923 staves,
F-clef staves twice) & training \\
CantusDB corpus & ten manuscripts; notes from
\citet{klugseder2026volpiano}, text from the Cantus Database & 10,463 &
training \\
Transkribus corpus & four Gothic manuscripts, note and text ground truth
& 611 & training \\
Graz manuscripts and prints & ten manuscripts and five printed chant
books, gold & 1,487 & training \\
14 further Graz volumes & gold & 463 & training \\
Silver, manuscripts & two-reader filter (Section 4.4) & 2,446 &
training \\
Silver, prints, capped & two-reader filter & 1,500 & training \\
Images with several staves & 5,585 sections, 529 half pages, 284 whole
pages; halves and pages three times & 8,024 & training \\
\textbf{Sum} & & \textbf{29,627} & \\
AMNLT validation & as published & 1,117 staves & checkpoint selection \\
Validation pages & whole pages with a target for every staff & 9 pages &
checkpoint selection \\
AMNLT test & as published: Solesmes 86, Einsiedeln 182, Salzinnes 294 &
562 staves & reported only \\
Held-out manuscripts & A-Wn 15484 (square); A-Geras 6, A-Gu 17, A-VOR
254 (German Gothic); A-KN 65 (Bohemian Gothic); notes only & 200 staves
(40 each) & reported only \\
Test pages; pages of the held-out manuscripts & & 14; 20 pages &
reported only \\
Pool & all Transkribus staves with note ground truth, 27 sources & 6,764
staves & reported only \\
Aldersbach check set & D-Mbs Clm 2766 (German Gothic), five pages not in
training, corrected by the author from a draft of P-d & 55 staves &
reported only \\
\end{longtable}

\subsection{AMNLT}\label{amnlt}

We use the three real datasets of AMNLT at staff level in the splits as
published: 5,602 staves after removal of duplicates (Einsiedeln 1,816,
Salzinnes 2,939, Solesmes 847; the AMNLT paper gives 2,965 and 854
\citep[Table 1]{fuentes2026amnlt}), of which 3,923 are for training,
1,117 for validation and 562 for testing. GregoSynth is not used. Two
properties of the data matter for everything that follows. The splits
are not page-disjoint: the 847 Solesmes images come from 75 pages, of
which 73 occur in more than one split, and all 176 test folios of
Salzinnes also occur in the training split. Figures on AMNLT validation
and test are therefore optimistic for our models and, by the same token,
for the published ones. Second, word boundaries are encoded only in
Solesmes; in Einsiedeln and Salzinnes a space precedes every syllable. A
model trained on that convention writes text syllable by syllable on
other sources too. From the fourth run on we therefore use a version in
which the syllables of Einsiedeln and Salzinnes are aligned with the
chant texts of the Cantus Database and spaces remain only before words
(3,822 of 4,755 lines rewritten).

\subsection{Chant aligned through the Cantus
Database}\label{chant-aligned-through-the-cantus-database}

The largest block rests on prior work of the project: ground truth that
links the Volpiano melodies of the Cantus Database to the staves of ten
manuscripts and records, for each staff, clef and notes as staff
positions \citep{klugseder2026volpiano}. That ground truth has no
lyrics. We add them from the full texts of the Cantus Database and take
word and syllable boundaries from the Volpiano string, where
\texttt{-\/-\/-} separates words, \texttt{-\/-} syllables and \texttt{-}
neumes inside a syllable. Where the count of words or syllables does not
match, the unresolved words are assigned by reading the image with a
general vision-language model under fixed checking rules; the author
checked two samples against the source images using visual review sheets
(25 of 25 and 19 of 19 sampled syllables correct). The result is a set
of 30,188 lines from eleven sources, split by page.

A staff enters the training corpus only if its image can be cut cleanly,
and that is the bottleneck. A crop must show the staff and its text line
and nothing of the next staff (Section 8.4). Two image filters, one for
a line of the neighbouring staff at the lower edge and one for cut text,
discard 62 \% of the candidates and leave 10,884 staves from ten
manuscripts (8,781 training, 1,060 validation, 1,043 test). The pages
are available at a width of 1,600 px only, so a staff step measures 9 to
13 px in the original; the crops are scaled up to 1,540 px (Section 5).
One of the ten manuscripts, CH-E 611, is the manuscript of AMNLT
Einsiedeln and is used for training only; the consequence for the
benchmark is discussed in Section 7.1.

\subsection{Ground truth from
Transkribus}\label{ground-truth-from-transkribus}

A second source of aligned training data is the corrected notation and
text exported from Transkribus: manuscripts and printed chant books
whose notes were transcribed there as a code of staff positions and
whose text was transcribed as corrected lines, with abbreviations
expanded and decorated initials often omitted. The Austrian sources were
largely digitised by the author; image access and rights are listed
under Data and code availability.

In an earlier project, chant books of this corpus were processed in
Transkribus. Student assistants transcribed the notation of about 1,500
pages by hand, and in 2024 the PRAIG group trained a CRNN on that ground
truth for use in Transkribus, within the project and not publicly; this
legacy model is no longer used and is not the second reader of Section
4.4. Both the students' ground truth and the legacy model's
transcriptions were corrected and reused, so part of the note ground
truth of this article goes back to corrected readings of an earlier
recogniser rather than to transcriptions from the image alone; for the
held-out manuscript A-KN 65 the project notes record this for pages 53
to 82.

The commas of the note code separate neumes, not syllables (they match
the number of syllables in only 8 to 43 \% of the lines, depending on
the source), so lines are paired directly where the counts agree and
otherwise by the image-reading step of Section 4.2 or the syllable gate
of Section 4.4. This gave 569 training staves from four Gothic
manuscripts, 1,769 lines from ten manuscripts and five printed chant
books of the Graz collection, and 515 lines of gold from 14 further Graz
volumes.

Building these corpora exposed errors in the joining of notes and text,
in the reading order of text regions, in the restoration of omitted
initials and in the syllabification. Two things follow for the method.
The model reproduces such errors on other staves with the same chant;
they do not average out. And an error in which a wrong syllable count is
compensated by a spare group of notes passes both the two-reader filter
and a visual check of the assignment; it was noticed on a review sheet
as an error of the text and measured by deriving the targets again under
a corrected rule. The corrections were written back to the source in
Transkribus, so that a new export does not bring the errors back, and
the rebuilt corpora entered training with P-e, which was not adopted as
the default model (Section 8.2).

\subsection{Silver staves from a two-reader
filter}\label{silver-staves-from-a-two-reader-filter}

Corrected text exists for more than 3,000 chant-book pages and note
ground truth for about 1,500 (Section 4.3). About 690 pages have both
and were converted to GABC by aligning syllables to neume groups; the
note ground truth of the remaining pages does not enter the training of
the model; it serves for evaluation and, in part, for training the
second reader described below. For the pages with corrected text but no
note ground truth, the notes come from readings, and silver targets are
built in two stages: agreement on notes and clef, followed by checks on
the alignment of syllables to neumes. In the first stage a staff is kept
only where two different recognisers read the same sequence of staff
positions and the same initial clef. The first reader is the page model
P-b. The second is the recogniser of the predecessor project, a hybrid
of a CRNN with a CTC head and a Transformer decoder that reads staff
positions and no text \citep{klugseder2026hybrid}; its architecture and
origin are given in Appendix D.

In the filter, the second reader was decoded greedily from its attention
head alone, without the CTC score, the melody model and the concordance
scoring that its inference package adds. The two readers differ in
architecture and in training target, but their training data overlap:
the second reader was trained, among other data, on checked ground truth
from 28 Transkribus sources, including corrected lines of the Aldersbach
manuscript, and on parts of Salzinnes and Solesmes. It may therefore
have seen sources that also occur in our evaluation splits. The readers
are not independent in a strict sense, and they share systematic errors,
as shown below.

\textbf{From notes to aligned targets.} A silver target takes its notes
and the places of the syllables from the reading of the first reader and
its text from the corrected transcription. A second gate decides whether
the text can be laid under the notes: the reading must have as many
syllables as the corrected transcription, with similar syllable texts;
each syllable must begin at a boundary between the neume groups of the
second reader; no bar may stand inside a word; and the two readings must
agree in the number of accidentals and in the clef changes. On 421 gold
staves outside the training data of the first reader, 242 pass, and in
98.8 \% of these every syllable begins on the same note as in the gold;
in the 195 of them where assigning the neume groups of the second reader
directly to the syllables gives the same starts, all 195 are right. Of
19,191 candidate staves, 7,551 targets result, 6,918 of them usable for
training (prints 4,757, Aldersbach 1,234, Graz manuscripts 927). With
their standardised printing types, printed chant books resemble the
individual letterforms of manuscripts only slightly and are therefore
capped at 1,500 examples per epoch. As training examples for the
alignment of syllables to neumes, however, they are valuable.

\textbf{Calibration.} Table 2 gives the calibration on staves with human
note ground truth. The staves come from pages that the index of the
second reader assigns to validation and test, that is, pages on which it
was not trained; its checkpoint, however, was selected on validation
lines, and other pages of the same manuscripts are in its training data.
The figures therefore describe agreement within known manuscripts, not
on unseen ones. On 1,474 manuscript staves of validation and test pages,
strict agreement passes 59.8 \% of the staves and lowers the pitch error
from 2.61 \% (page model alone) to 0.18 \%; on printed chant books it
passes 80.6 \% at 0.21 \%. A first calibration with a second reader that
did not fit the material passed only 19.9 \%; the yield of the filter
depends on the second reader. The target precision of 99 \% is not
demonstrated: 95.8 \% of the passed manuscript staves are exactly right
when measured against the human ground truth. On the test pages alone
(773 staves) strict agreement passes 59.3 \% at a precision of 95.6 \%
and a pitch SER of 0.16 \%.

\needspace{14\baselineskip}

\begin{longtable}[]{@{}
  >{\raggedright\arraybackslash}p{(\linewidth - 14\tabcolsep) * \real{0.1644}}
  >{\raggedright\arraybackslash}p{(\linewidth - 14\tabcolsep) * \real{0.1781}}
  >{\raggedleft\arraybackslash}p{(\linewidth - 14\tabcolsep) * \real{0.1027}}
  >{\raggedleft\arraybackslash}p{(\linewidth - 14\tabcolsep) * \real{0.1027}}
  >{\raggedleft\arraybackslash}p{(\linewidth - 14\tabcolsep) * \real{0.0959}}
  >{\raggedleft\arraybackslash}p{(\linewidth - 14\tabcolsep) * \real{0.1438}}
  >{\raggedleft\arraybackslash}p{(\linewidth - 14\tabcolsep) * \real{0.0959}}
  >{\raggedleft\arraybackslash}p{(\linewidth - 14\tabcolsep) * \real{0.1164}}@{}}
\caption{Two-reader filter, calibration on validation and test pages
with human note ground truth. Precision: share of passed staves whose
pitch sequence and clef equal the ground truth.}\tabularnewline
\toprule\noalign{}
\begin{minipage}[b]{\linewidth}\raggedright
Material
\end{minipage} & \begin{minipage}[b]{\linewidth}\raggedright
Stage
\end{minipage} & \begin{minipage}[b]{\linewidth}\raggedleft
Staves
\end{minipage} & \begin{minipage}[b]{\linewidth}\raggedleft
Passed
\end{minipage} & \begin{minipage}[b]{\linewidth}\raggedleft
Yield
\end{minipage} & \begin{minipage}[b]{\linewidth}\raggedleft
Precision
\end{minipage} & \begin{minipage}[b]{\linewidth}\raggedleft
Pitch SER
\end{minipage} & \begin{minipage}[b]{\linewidth}\raggedleft
Initial clef
\end{minipage} \\
\midrule\noalign{}
\endfirsthead
\toprule\noalign{}
\begin{minipage}[b]{\linewidth}\raggedright
Material
\end{minipage} & \begin{minipage}[b]{\linewidth}\raggedright
Stage
\end{minipage} & \begin{minipage}[b]{\linewidth}\raggedleft
Staves
\end{minipage} & \begin{minipage}[b]{\linewidth}\raggedleft
Passed
\end{minipage} & \begin{minipage}[b]{\linewidth}\raggedleft
Yield
\end{minipage} & \begin{minipage}[b]{\linewidth}\raggedleft
Precision
\end{minipage} & \begin{minipage}[b]{\linewidth}\raggedleft
Pitch SER
\end{minipage} & \begin{minipage}[b]{\linewidth}\raggedleft
Initial clef
\end{minipage} \\
\midrule\noalign{}
\endhead
\bottomrule\noalign{}
\endlastfoot
Manuscripts & page model P-b alone & 1,474 & 1,474 & 100 \% & 70.2 \% &
2.61 \% & 91.4 \% \\
Manuscripts & second reader alone & 1,474 & 1,474 & 100 \% & 74.1 \% &
1.59 \% & 97.5 \% \\
Manuscripts & strict agreement & 1,474 & 881 & 59.8 \% & 95.8 \% & 0.18
\% & 99.4 \% \\
Manuscripts & at most one note differs & 1,474 & 1,126 & 76.4 \% & 86.4
\% & 0.63 \% & 99.6 \% \\
Prints & page model P-b alone & 165 & 165 & 100 \% & 80.0 \% & 1.58 \% &
100 \% \\
Prints & second reader alone & 165 & 165 & 100 \% & 93.9 \% & 0.60 \% &
100 \% \\
Prints & strict agreement & 165 & 133 & 80.6 \% & 97.0 \% & 0.21 \% &
100 \% \\
\end{longtable}

\textbf{The filter also finds errors in the human ground truth.} Part of
the missing precision is not an error of the readers. On the gold of the
14 Graz volumes, the two readers agree strictly on 927 staves, and on
103 of them the human ground truth differs. Of the 78 of these staves
checked against the source images, 68 favour the readers and 8 the
ground truth (details in Appendix D); the ground truth was corrected in
95 staves. The 8 cases in which the ground truth is right show where
agreement is weak evidence: both readers take the same foreign object
for a note (a custos, the small guide sign at the end of a line that
anticipates the first note of the next; a letter; a note shining through
from the verso), or both read an F clef as a C clef. In this inspection,
agreement of two different recognisers was reliable for substituted
notes, but did not rule out shared extra notes or F clefs read as C
clefs.

\subsection{Evaluation sets}\label{evaluation-sets}

Checkpoint selection (Section 5) uses only AMNLT validation and, for the
page models, the nine validation pages. The other sets are used for
evaluation and error analysis; observations on them informed later
choices of training data, such as counting F-clef staves twice (Section
8.1), but never the choice of a checkpoint. We distinguish unseen
manuscripts, unseen pages of manuscripts that occur in training, and the
wider Transkribus pool. The held-out manuscripts are unseen: 200 staves
with verified notes, 40 from each of five manuscripts that occur in no
training list, and 20 of their pages on which every staff has a target.
Their note targets carry no lyrics; for lyrics evaluation we separately
use the corrected Transkribus transcriptions, available for 154 to 162
of the staves depending on the comparison. The 9 validation and 14 test
pages are unseen pages of known manuscripts: every staff on them has a
target, and they come almost entirely from three manuscripts of the
chant corpus with small notation. The Aldersbach check set below is of
the same kind. The pool extends the staff measurement to all 6,764
Transkribus staves with note ground truth, including the training
sources.

The Aldersbach check set was built for one question, dropped notes at
abbreviated text (Section 8.2). The manuscript, D-Mbs Clm 2766, is not
unseen: 1,314 of its staves are silver staves in the training data of
P-d.~The five pages of the set are among the 50 pages of the manuscript
that occur in no training list. The author corrected notes and text of
their 55 staves in a purpose-built editor, starting from a draft read by
P-d, with marks where the second reader disagrees. The set therefore has
a known bias: an error of the draft that the corrector did not notice
counts as correct.

\section{Model and Training}\label{model-and-training}

\textbf{Model.} LightOnOCR-3-1B consists of a Pixtral vision transformer
(24 layers, patches of 14 px, images up to 1,540 px on the longest
edge), a projector that merges 2 \ensuremath{\times} 2 patches into one image token, so
that one image token covers 28 px, and a Qwen3 decoder with 28 layers
\citep{lightonocr3card, lightonocr2026, agrawal2024pixtral, yang2025qwen3}.
According to \citet{lightonocr2026}, the vision encoder is initialised
from Mistral-Small-3.1 and the decoder from a pre-trained Qwen3 model.
We change nothing in the architecture and add no vocabulary; the prefix
\texttt{\textless{}m\textgreater{}} is split by the tokenizer of the
decoder like any other text. All weights are trained, with a learning
rate of 1e-4 for the vision encoder and projector and 5e-5 for the
decoder.

\textbf{Resolution and the strip layout.} The quantity that decides
whether pitches can be read is the height of a staff step, half the
distance between two staff lines, relative to the patch. A staff of
AMNLT is very wide (the Einsiedeln staves measure about 3,300 by 350
px); scaled to 1,540 px, a staff step falls to a median of 13.8 px for
Einsiedeln, the size of one patch. We therefore cut a long staff at
columns with little ink into two to four pieces and stack them, without
overlap and without separate targets for the pieces. This strip layout
roughly doubles the vertical resolution (29 px per staff step for
Einsiedeln) at the price of more image tokens. Small crops are scaled up
to 1,540 px. All runs use this layout. A control run with unstacked
staves was proposed and never run.

\textbf{Images with several staves.} The page models are trained with
the same code on additional examples that show more than one staff:
sections of two to four consecutive staves, half pages and whole pages,
cut between staves below the text line. The target is the sequence of
the staff targets with a line break after each. Such examples can only
be built from pages on which every staff has a target; a check that
looks for staves missing from the list (a four-line comb and jumps in
the line spacing) removed 94 pages. In the default model, half and whole
pages are counted three times per epoch; examples with several staves
make up 27 \% of the examples and 62 \% of the target characters.

\textbf{Training.} One example per forward pass with gradient
accumulation over 8, bf16 autocast, gradient checkpointing, AdamW
without weight decay, 50 warm-up steps, cosine decay to zero, gradient
clipping at 0.5, seed 0, four epochs, one H100. The default model took
13.4 h for 4 \ensuremath{\times} 29,627 examples (0.41 s per example, 26.8 GB peak). Every
run starts from the published weights, except P-ft, which continues the
last staff model for one epoch at half the learning rates. Staves whose
target begins with an F clef are counted twice from P-c on, to counter
the weakness described in Section 8.1.

\textbf{Weights.} The decoder weights of the published checkpoint were
renamed to match the library version we used, and their identity was
verified tensor by tensor before training (Appendix A).

\textbf{Selection.} A checkpoint is written after every epoch. For the
staff models the epoch with the lowest pitch SER on AMNLT validation is
taken. For the page models two criteria are used, pitch SER on AMNLT
validation and on the nine validation pages: when they favour different
epochs and these differ by less than 0.1 percentage points on AMNLT
validation, the validation pages decide; otherwise AMNLT validation
decides. The held-out sets never enter the choice, which has a price:
for P-b the third epoch read the held-out manuscripts better than the
chosen fourth (1.46 against 1.74 \%).

\textbf{Decoding.} Greedy, without repetition penalty, because repeated
notes are regular in chant; loops are detected afterwards (the same
sequence at least six times at the end). The output limit is 4,096
tokens for pages.

\section{Experiments}\label{experiments}

\textbf{Runs.} Table 3 lists the runs; Table 9 in Appendix E gives them
in full, with the internal names of the repository, the chosen epoch,
the initial clef and the text CER. Each run changes the training lists
and nothing in architecture or settings (P-ft aside): the staff models
add sources step by step, the page models add images with several staves
and then larger datasets with corrected targets. Several things often
change together in one step (sources, amount, text convention,
weighting, image forms), and each step is one run; the series shows a
course, not isolated effects. We also trained LightOnOCR-3-4B on the
data of the last staff model (Section 7.5).

\textbf{Measures.} \emph{Pitch SER} (pitch symbol error rate), our main
measure, is the edit distance over the position letters \texttt{a} to
\texttt{m}, without clefs, custodes, accidentals and shape signs,
divided by the number of reference notes. It measures edits to
staff-position letters, not errors in absolute musical pitch, and it is
defined for every source, including those with notes only. \emph{Initial
clef} is the share of staves whose first clef is right. \emph{MER}
(music error rate) is the edit distance over all music characters,
\emph{CER} (character error rate) over the characters of the lyrics
without spaces, \emph{SylER} (syllable error rate) over syllables.
\emph{AMLER} (aligned music and lyrics error rate), \emph{bWER} and
\emph{AlER} (alignment error rate) follow \citet[Section
6]{fuentes2026amnlt} and \citet{vidal2023bwer}: AMLER is the normalised
edit distance on the joint sequence of syllables and music tokens,
averaged over examples; bWER is the same without regard to order; AlER =
(AMLER \ensuremath{-} bWER) / AMLER. Because AlER is a share and rises when the
content error falls, we report AMLER \ensuremath{-} bWER in percentage points beside
it. Where the AMNLT study is silent on tokenisation we take syllables,
single parentheses and single music characters as tokens.

\textbf{Error kinds and page scoring.} From Section 8 on, the pitch
error is split into missing, extra and wrong notes, and we count
\emph{dropped units}: syllables whose text and all of whose notes are
deleted in the alignment, a strict count and therefore a lower bound.
For a page, the pitch SER is computed over the page as one sequence,
ignoring line breaks, and a staff counts as found when at least half of
its notes are matched in that alignment. Where a page reading is scored
against the targets of single staves (Section 8.3), its lines are first
aligned to the staves.

\textbf{Procedure.} Training and inference ran on one H100 of the MUSICA
cluster; every checkpoint was evaluated on all validation sets and on
the held-out manuscripts in the same job. The test split of AMNLT was
read once per model, after the choice of the checkpoint. Each run has
one seed; we have no estimate of the variance between runs. Between
epochs of one run, values on the held-out manuscripts move by about one
point for the early runs and by a few tenths for the later ones.

\needspace{14\baselineskip}

\begin{longtable}[]{@{}
  >{\raggedright\arraybackslash}p{(\linewidth - 6\tabcolsep) * \real{0.1667}}
  >{\raggedright\arraybackslash}p{(\linewidth - 6\tabcolsep) * \real{0.5439}}
  >{\raggedleft\arraybackslash}p{(\linewidth - 6\tabcolsep) * \real{0.1228}}
  >{\raggedleft\arraybackslash}p{(\linewidth - 6\tabcolsep) * \real{0.1667}}@{}}
\caption{Runs. Pitch SER in \% on AMNLT validation and on the 200
held-out staves; the full table with internal names, chosen epoch,
initial clef and text CER is Table 9 in Appendix E. P-e was not adopted
(Section 8.2).}\tabularnewline
\toprule\noalign{}
\begin{minipage}[b]{\linewidth}\raggedright
Name
\end{minipage} & \begin{minipage}[b]{\linewidth}\raggedright
Training data (examples per epoch)
\end{minipage} & \begin{minipage}[b]{\linewidth}\raggedleft
AMNLT val
\end{minipage} & \begin{minipage}[b]{\linewidth}\raggedleft
Held-out
\end{minipage} \\
\midrule\noalign{}
\endfirsthead
\toprule\noalign{}
\begin{minipage}[b]{\linewidth}\raggedright
Name
\end{minipage} & \begin{minipage}[b]{\linewidth}\raggedright
Training data (examples per epoch)
\end{minipage} & \begin{minipage}[b]{\linewidth}\raggedleft
AMNLT val
\end{minipage} & \begin{minipage}[b]{\linewidth}\raggedleft
Held-out
\end{minipage} \\
\midrule\noalign{}
\endhead
\bottomrule\noalign{}
\endlastfoot
S-AMNLT & AMNLT (3,923) & 1.77 & 24.92 \\
S+C3 & + 2,160 CantusDB staves, three manuscripts (6,083) & 1.79 &
9.55 \\
S+C10 & 8,781 CantusDB staves, ten manuscripts, in place of the 2,160
(12,704) & 1.86 & 5.84 \\
S+C10w & as before, AMNLT with word boundaries (12,704) & 1.68 & 4.67 \\
S+C10w+T & + 569 staves, four Gothic manuscripts (13,273) & 1.80 &
2.11 \\
P-ft & previous model continued, + 2,390 multi-staff images (15,663) &
1.89 & 2.28 \\
P-b & + 4,551 multi-staff images (17,824) & 1.71 & 1.74 \\
P-c & cleaned Transkribus targets, Graz corpus, F-clef staves twice
(22,737) & 1.77 & 1.27 \\
P-d & + gold of 14 Graz volumes, silver, pages three times (29,627) &
1.74 & 1.31 \\
P-e & + text-dropout examples (1,598), rebuilt Transkribus corpora
(31,393) & 1.75 & 1.93 \\
4B & LightOnOCR-3-4B, data of S+C10w+T (13,273) & 1.80 & 2.38 \\
\end{longtable}

\section{Results}\label{results}

\subsection{The AMNLT test split}\label{the-amnlt-test-split}

Table 4 and Figure 3 compare our models with the published values on the
test split of the three real datasets. The first run, S-AMNLT, was
trained on the published AMNLT training split and nothing else (3,923
staves after local removal of duplicates; we have not verified by
identifiers that this split equals the one used in the AMNLT study). Its
AMLER is 9.92 \% on Solesmes, 2.25 \% on Einsiedeln and 5.31 \% on
Salzinnes; the best published values are 20.98 \% (unfolding CRNN), 5.14
\% (DAN) and 7.32 \% (SMT). The default model P-d reaches 8.80, 1.87 and
4.36 \%. Over all 562 staves P-d has a pitch SER of 1.25 \%, a lyrics
CER of 1.86 \% and an AMLER of 4.23 \%, with no truncated output and no
loop.

Splitting AMLER into bWER and the remainder shows where the gain lies.
On Solesmes the content error falls from 13.01 to 5.61 points and the
remainder from 7.97 to 3.19; on Einsiedeln from 4.27 to 1.10 and from
0.87 to 0.77; on Salzinnes from 6.30 to 3.53 and from 1.02 to 0.83 (best
published method against P-d). The model mainly recognises notes and
text more accurately; the alignment gains less, and AlER, a share of a
now smaller total, even rises for Einsiedeln.

The comparison has five limitations.

\begin{enumerate}
\def\labelenumi{\arabic{enumi}.}
\tightlist
\item
  \emph{Like for like.} Only S-AMNLT is restricted to the published
  training split. All later models have seen further chant.
\item
  \emph{Encoding and denominator.} Only Solesmes is directly comparable:
  the AMNLT study also trains and measures it in music-aware gabc, and
  our tokenisation matches the authors' as far as we know it. For
  Einsiedeln and Salzinnes the AMNLT study trains and measures on
  pseudo-gabc and splits it into multi-character tokens of a vocabulary;
  we train and measure on the common encoding with one token per music
  character. Our denominator is larger, and a clef error counts once for
  us and on every note there. Part of the distance on these two datasets
  may come from the counting. The same holds for MER.
\item
  \emph{Overlap.} From S+C3 on, the chant corpus contains CH-E 611, the
  manuscript of Einsiedeln, in training. The Einsiedeln figures of those
  models are not independent.
\item
  \emph{Splits.} The AMNLT splits are not page-disjoint. This favours
  our models and the published ones alike.
\item
  \emph{One seed.} Each of our figures comes from one run.
\end{enumerate}

The bWER of the published rows is not printed in the AMNLT study; we
computed it as AMLER \ensuremath{\cdot} (1 \ensuremath{-} AlER) from AlER values given to two
decimals, so it carries a rounding uncertainty of a few tenths of a
point.

By design, our approach belongs to the language-modelling family of the
AMNLT study (SMT, DAN, PaliGemma 2): models that turn the image into a
sequence without unfolding lines. DAN is therefore given as a separate
row. Shape-sign errors account for much of the remaining MER, so MER and
pitch SER capture different aspects of performance (Section 3): on
Salzinnes our MER is 8.3 \%, the pitch SER 1.4 \% (Section 8.6).

\needspace{15\baselineskip}

\begin{longtable}[]{@{}
  >{\raggedright\arraybackslash}p{(\linewidth - 10\tabcolsep) * \real{0.2000}}
  >{\raggedright\arraybackslash}p{(\linewidth - 10\tabcolsep) * \real{0.3333}}
  >{\raggedleft\arraybackslash}p{(\linewidth - 10\tabcolsep) * \real{0.1167}}
  >{\raggedleft\arraybackslash}p{(\linewidth - 10\tabcolsep) * \real{0.1167}}
  >{\raggedleft\arraybackslash}p{(\linewidth - 10\tabcolsep) * \real{0.1167}}
  >{\raggedleft\arraybackslash}p{(\linewidth - 10\tabcolsep) * \real{0.1167}}@{}}
\caption{AMNLT test split (562 staves), in \%. Published values from
Tables 2 and 3 of \citet{fuentes2026amnlt}; their bWER is derived as
described in the text. *Einsiedeln and Salzinnes: different encoding and
denominator, see reservation 2. MER, CER, SylER and AlER are given in
Table 10, Appendix F.}\tabularnewline
\toprule\noalign{}
\begin{minipage}[b]{\linewidth}\raggedright
Dataset
\end{minipage} & \begin{minipage}[b]{\linewidth}\raggedright
Model
\end{minipage} & \begin{minipage}[b]{\linewidth}\raggedleft
Pitch SER
\end{minipage} & \begin{minipage}[b]{\linewidth}\raggedleft
AMLER
\end{minipage} & \begin{minipage}[b]{\linewidth}\raggedleft
bWER
\end{minipage} & \begin{minipage}[b]{\linewidth}\raggedleft
AMLER \ensuremath{-} bWER
\end{minipage} \\
\midrule\noalign{}
\endfirsthead
\toprule\noalign{}
\begin{minipage}[b]{\linewidth}\raggedright
Dataset
\end{minipage} & \begin{minipage}[b]{\linewidth}\raggedright
Model
\end{minipage} & \begin{minipage}[b]{\linewidth}\raggedleft
Pitch SER
\end{minipage} & \begin{minipage}[b]{\linewidth}\raggedleft
AMLER
\end{minipage} & \begin{minipage}[b]{\linewidth}\raggedleft
bWER
\end{minipage} & \begin{minipage}[b]{\linewidth}\raggedleft
AMLER \ensuremath{-} bWER
\end{minipage} \\
\midrule\noalign{}
\endhead
\bottomrule\noalign{}
\endlastfoot
Solesmes & published, best AMLER (unfolding CRNN) & -- & 20.98 & 13.01 &
7.97 \\
Solesmes & published, DAN & -- & 23.17 & 14.13 & 9.04 \\
Solesmes & S-AMNLT & 1.87 & 9.92 & 6.64 & 3.28 \\
Solesmes & S+C10w+T & 1.50 & 9.35 & 5.91 & 3.44 \\
Solesmes & P-d & 1.23 & 8.80 & 5.61 & 3.19 \\
Einsiedeln* & published, best AMLER (DAN) & -- & 5.14 & 4.27 & 0.87 \\
Einsiedeln* & S-AMNLT & 1.23 & 2.25 & 1.56 & 0.69 \\
Einsiedeln* & S+C10w+T & 1.00 & 1.84 & 1.19 & 0.65 \\
Einsiedeln* & P-d & 1.10 & 1.87 & 1.10 & 0.77 \\
Salzinnes* & published, best AMLER (SMT) & -- & 7.32 & 6.30 & 1.02 \\
Salzinnes* & published, DAN & -- & 18.73 & 14.05 & 4.68 \\
Salzinnes* & S-AMNLT & 1.63 & 5.31 & 4.47 & 0.84 \\
Salzinnes* & S+C10w+T & 1.32 & 4.54 & 3.86 & 0.68 \\
Salzinnes* & P-d & 1.41 & 4.36 & 3.53 & 0.83 \\
\end{longtable}

\begin{figure}
\centering
\pandocbounded{\includegraphics[keepaspectratio,alt={AMLER on the AMNLT
test split, divided into the content error (bWER) and the remainder
attributed to alignment. Light bars: published values; dark bars: this
work. S-AMNLT is trained on the AMNLT training data only. *Different
encoding and denominator, not directly comparable. The bWER of the
published values is derived from AMLER and
AlER.}]{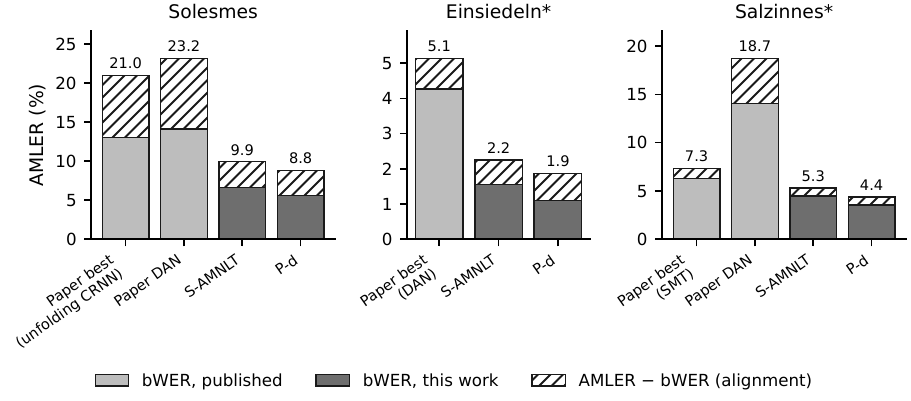}}
\caption{AMLER on the AMNLT test split, divided into the content error
(bWER) and the remainder attributed to alignment. Light bars: published
values; dark bars: this work. S-AMNLT is trained on the AMNLT training
data only. *Different encoding and denominator, not directly comparable.
The bWER of the published values is derived from AMLER and AlER.}
\end{figure}

\subsection{The benchmark is flat, generalisation is
not}\label{the-benchmark-is-flat-generalisation-is-not}

Figure 4 shows the two curves that shaped the project. Across the nine
runs from S-AMNLT to P-d, the pitch SER on AMNLT validation lies between
1.68 and 1.89 \%; no addition of data moves it, and the later
exploratory run P-e lies in the same range (1.75 \%). On the held-out
manuscripts it falls from 24.92 \% to 1.31 \%. A model selected and
reported on AMNLT alone would have looked finished after the first run.

AMNLT validation holds out staves, not pages or manuscripts (Section
4.1); it measures the reading of further staves from the same books. The
fall on the held-out manuscripts proceeds in steps, each of which
coincides with an addition to the training data (Table 3, Figure 4);
with one run per step these are observations, not isolated effects. Two
steps stand out. Chant from the CantusDB corpus takes the error from
24.9 to 9.6 \% with 2,160 staves from three manuscripts and to 5.8 \%
with 8,781 staves from ten. Later, adding 569 staves from four Gothic
manuscripts reduces the error from 4.7 to 2.1 \%. This gain is
concentrated in A-Gu 17 and A-VOR 254, which use the same German Gothic
notation as the new training sources (Table 7), suggesting that coverage
of the target notation matters alongside corpus size.

A wider check confirms the picture for the last staff model: on 3,015
staves from 17 further Transkribus sources that are in no training set
it has a pitch SER of 2.62 \%, on the four training sources 1.06 \%. One
further source in lesson tone is left out, because its ground truth
counts a long recitation bar as one note while the model writes one note
per syllable; the author confirmed against the source images that this
is a difference of convention. For comparison,
\texttt{hybrid\_hybrid\_lang}, the earlier checkpoint of the predecessor
project's recogniser from which the second reader of Section 4.4 was
later fine-tuned, trained on about 38,000 staves of the chant ground
truth only and decoded greedily from its attention head, without melody
model and concordances, reads the same 200 held-out staves at 13.88 \%,
against 9.55 \% for S+C3 with 6,083 training staves.

\begin{figure}
\centering
\pandocbounded{\includegraphics[keepaspectratio,alt={Pitch SER on AMNLT
validation and on the held-out manuscripts over the runs with the 1B
model. P-ft continues the previous model; all other runs start from the
published weights. P-e was not adopted (Section
8.2).}]{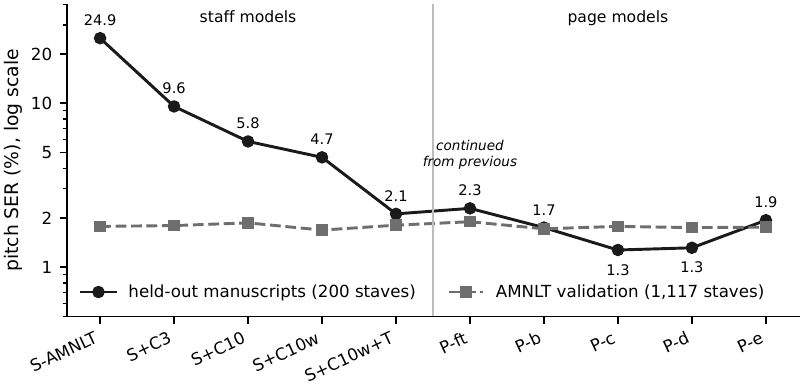}}
\caption{Pitch SER on AMNLT validation and on the held-out manuscripts
over the runs with the 1B model. P-ft continues the previous model; all
other runs start from the published weights. P-e was not adopted
(Section 8.2).}
\end{figure}

\subsection{Whole pages}\label{whole-pages}

\textbf{Without page training the ability is absent.} The last staff
model, given a whole page, never writes a line break; it reads about one
staff and stops. On the 23 validation and test pages its pitch SER is
above 93 \%, on the pages of the held-out manuscripts 65.5 \%.

\textbf{It is learned quickly.} One further epoch with 13 \% of images
with several staves (P-ft) brings the validation and test pages to about
10 \% and the held-out pages to 2.4 \%, equal to cropped staves of the
same pages. Training from the published weights with 26 \% such images
(P-b) reaches 2.9 \% on the nine validation pages and 2.1 \% on the
held-out pages and at the same time improves staff reading (held-out
staves 2.11 to 1.74 \%). The page models recovered 98 to 100 \% of the
staves and produced no truncated or looping output on pages. Table 5
gives the three page models on all sets.

\needspace{14\baselineskip}

\begin{longtable}[]{@{}
  >{\raggedright\arraybackslash}p{(\linewidth - 6\tabcolsep) * \real{0.5500}}
  >{\raggedleft\arraybackslash}p{(\linewidth - 6\tabcolsep) * \real{0.1500}}
  >{\raggedleft\arraybackslash}p{(\linewidth - 6\tabcolsep) * \real{0.1500}}
  >{\raggedleft\arraybackslash}p{(\linewidth - 6\tabcolsep) * \real{0.1500}}@{}}
\caption{Page models, chosen epoch 4 each. Pitch SER in \% unless
stated. ``Without GT errors'': without two staves of one validation page
on which the image shows that the model is right and the target
wrong.}\tabularnewline
\toprule\noalign{}
\begin{minipage}[b]{\linewidth}\raggedright
Measure
\end{minipage} & \begin{minipage}[b]{\linewidth}\raggedleft
P-b
\end{minipage} & \begin{minipage}[b]{\linewidth}\raggedleft
P-c
\end{minipage} & \begin{minipage}[b]{\linewidth}\raggedleft
P-d
\end{minipage} \\
\midrule\noalign{}
\endfirsthead
\toprule\noalign{}
\begin{minipage}[b]{\linewidth}\raggedright
Measure
\end{minipage} & \begin{minipage}[b]{\linewidth}\raggedleft
P-b
\end{minipage} & \begin{minipage}[b]{\linewidth}\raggedleft
P-c
\end{minipage} & \begin{minipage}[b]{\linewidth}\raggedleft
P-d
\end{minipage} \\
\midrule\noalign{}
\endhead
\bottomrule\noalign{}
\endlastfoot
AMNLT validation, pitch SER & 1.71 & 1.77 & 1.74 \\
AMNLT validation, AMLER & 4.45 & 4.57 & 4.30 \\
CantusDB validation & 5.66 & 6.25 & 5.72 \\
Transkribus validation & 0.42 & 0.46 & 0.54 \\
Transkribus validation, lyrics CER & 5.50 & 3.41 & 3.30 \\
Held-out manuscripts, 200 staves & 1.74 & 1.27 & 1.31 \\
Held-out manuscripts, initial F clef & 24 of 47 & 32 of 47 & 30 of 47 \\
Held-out manuscripts, initial C clef & 150 of 153 & 151 of 153 & 152 of
153 \\
Whole pages, validation (9) & 2.94 & 3.98 & 3.34 \\
Whole pages, validation, without GT errors & 2.97 & 3.72 & 3.12 \\
Whole pages, test (14) & 5.39 & 4.06 & 3.82 \\
Whole pages, held-out manuscripts (20) & 2.1 & 1.9 & 1.9 \\
\end{longtable}

The differences between the three are small against what nine and
fourteen pages can resolve: P-c is worse than P-b on the validation
pages and better on the test pages. The changes are consistent with what
they were aimed at, although none of these effects was isolated
experimentally. Counting F-clef staves twice and adding the Graz corpus
coincided with a rise of the initial F clefs on the held-out manuscripts
from 24 to 32 of 47 and a fall of the pitch SER there from 1.74 to 1.27
\%; the cleaned Transkribus targets coincided with a fall of the lyrics
CER on Transkribus validation from 5.50 to 3.41 \%.

\textbf{Resolution decides between page and parts.} Figure 5 puts two
things side by side. On a whole page scaled to 1,540 px, a staff step
measures 7 to 13 px in the sources we could measure, below the 14 px of
a patch everywhere and about half of it for the three manuscripts of the
chant corpus that make up most of the validation and test pages. Read as
a whole page, those pages have a pitch SER of 3.6 \%, as two halves 2.3
\%, as sections of a few staves 2.6 \%, as single staves cut from known
boxes 1.7 \%. On the pages of the held-out manuscripts, with larger
notation, the order is reversed: the whole page (1.9 \%) and its parts
(1.2 \%) read better than the single staves (2.5 \%). The figures for
single staves use boxes known from the ground truth; in use they would
need a staff detector, whose errors are not included.

A larger input edge without training does not help. The page model P-ft
read with 1,540, 2,300 and 2,800 px gives 9.5, 13.7 and 43.9 \% on the
validation and test pages: the model knows positions up to 110 patches
and the scale of the 1,540 px images, and nothing beyond. Reading time
on the H100 depends almost only on the length of the output: in a single
process, P-d needed 16.7 ms per output token and 38.4 s per validation
page on average (2,308 output tokens; first page left out as warm-up).

\begin{figure}
\centering
\pandocbounded{\includegraphics[keepaspectratio,alt={(a) Pixels per
staff step after scaling to 1,540 px, for whole pages, half pages and
staff crops of eight sources; the dotted line is the patch size. (b)
Pitch SER of P-d by reading form. The validation and test pages come
mostly from the three sources at the top of
(a).}]{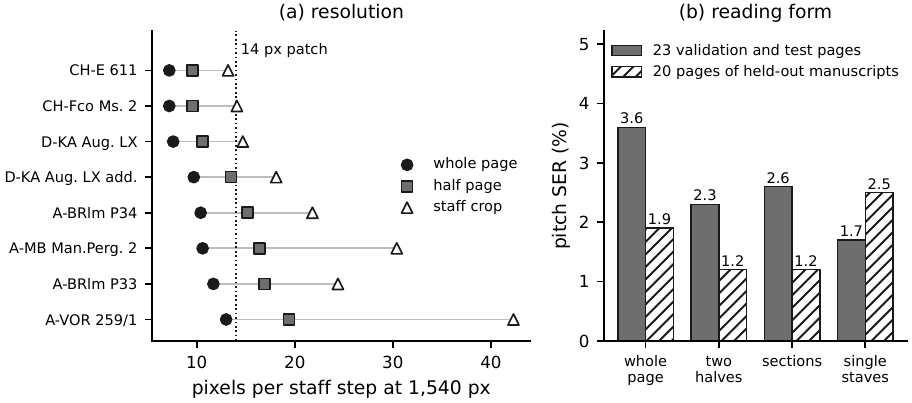}}
\caption{(a) Pixels per staff step after scaling to 1,540 px, for whole
pages, half pages and staff crops of eight sources; the dotted line is
the patch size. (b) Pitch SER of P-d by reading form. The validation and
test pages come mostly from the three sources at the top of (a).}
\end{figure}

\subsection{Choosing the checkpoint}\label{choosing-the-checkpoint}

The validation loss is not a guide to the best epoch (Figure 6). In
every run of four epochs it is lowest in epoch 2 or 3 and rises in epoch
4, while the page measures keep improving until epoch 4. P-e is the
exception: its best page result is in epoch 3 (3.58 against 3.60 \% in
epoch 4). For P-c the loss would have chosen epoch 2, with a pitch SER
of 8.4 \% on the validation pages, rather than epoch 4, with 4.0 \%; for
P-d it would have chosen epoch 3 (3.9 against 3.3 \%). All epochs are
listed in Table 8 in the appendix. The loss is computed with teacher
forcing on single staves of AMNLT; reading a page of two to three
thousand tokens without error is a different skill, and for P-b, P-c and
P-d the page curve is not yet flat after four epochs.

\begin{figure}
\centering
\pandocbounded{\includegraphics[keepaspectratio,alt={Validation loss and
reading measures over the epochs of the four page models. The ring marks
the epoch with the lowest loss; the chosen epoch is 4 for P-b, P-c and
P-d and 3 for P-e, which was not adopted (Section
8.2).}]{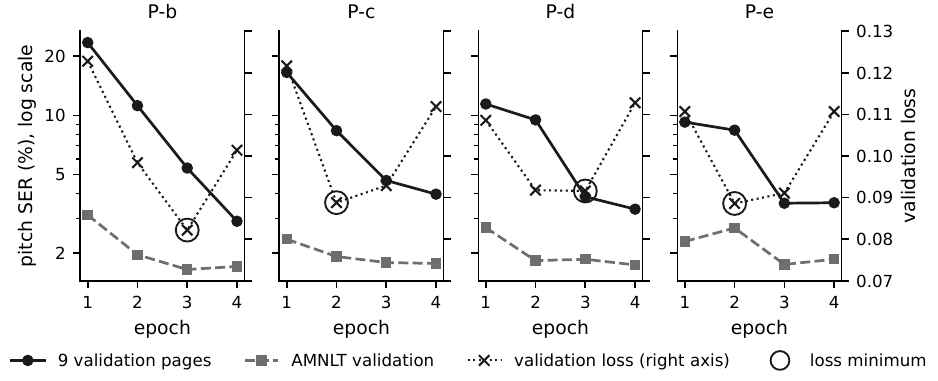}}
\caption{Validation loss and reading measures over the epochs of the
four page models. The ring marks the epoch with the lowest loss; the
chosen epoch is 4 for P-b, P-c and P-d and 3 for P-e, which was not
adopted (Section 8.2).}
\end{figure}

\subsection{A larger model}\label{a-larger-model}

LightOnOCR-3-4B is not an enlarged 1B but a different architecture with
about 4.5 billion parameters. With the data of the last staff model it
is level on AMNLT validation (1.80 \%) and slightly worse on the
held-out manuscripts (pitch SER 2.38 against 2.11 \%, text CER 8.14
against 7.90 \%). It needs more than three times the GPU memory (78.7
against 24 GB), reads about 1.75 times more slowly and produced more
loops in its early epochs. Its errors in the text are of a different
kind: it leaves out more and invents less, and it leaves abbreviations
as they stand in the image. We did not pursue it.

\section{Error Analysis}\label{error-analysis}

\subsection{The Gothic F clef}\label{the-gothic-f-clef}

The weakest point of every model in this article is the F clef of the
Gothic manuscripts. On the held-out staves P-d reads 152 of 153 initial
C clefs correctly and 30 of 47 initial F clefs. The typical error is an
F clef read as a C clef on the same line, usually with the notes that
follow written consistently for the wrong clef. The weakness is not
specific to this model. The recogniser of the predecessor project
confused F and C clefs on Salzinnes; in the pool measurement, 66 of 75
clef deviations in A-Wn 15484 and 32 of 39 in A-VOR 254 are of this
kind; and F read as C is the most frequent error shared by the two
readers of the filter (Section 4.4). The first page model made it worse:
its additional multi-staff lines begin with a C clef in 88 \% of the
cases, the share of F in training fell from about 17 to 14 \%, and 15
staves flipped from right to wrong against 4 in the other direction
(sign test p = 0.02). Counting F-clef staves twice recovered part of the
loss (24 to 32 of 47) and did not solve it. In the position encoding the
damage is confined to the clef token; for anyone who needs pitches, a
staff with a wrong clef is wrong throughout.

\subsection{Notes fall with their
syllable}\label{notes-fall-with-their-syllable}

The author, reading with the model manuscripts that it had not seen,
noticed that where a syllable of the text is missing in the reading, the
notes above it are often missing too, although they are clearly legible,
and that the words concerned are mostly abbreviated. This is an error of
the reading and follows from its form. The model writes a syllable and
its notes as one unit, the syllable first; when it cannot read the
syllable it tends to skip the unit as a whole. Abbreviations add a
second difficulty: the target expands the word, so one written sign
stands for several syllables, each with its own notes.

We therefore split the pitch error by kind (Figure 7). On the AMNLT test
split the three kinds are balanced and small. On the held-out
manuscripts, missing notes are the most frequent kind for P-d, 0.72 of
1.31 points, and they rose from P-c (0.49) while extra and wrong notes
fell. Eight of the eleven additional missing notes lie in one staff of
A-Wn 15484, in which the notes of a psalm differentia (the closing
formula of a psalm tone) stand above the abbreviation ``ps. Mag.''; P-d
writes the text and no notes, where P-c had invented the syllables
e-u-o-u-a-e to carry them. On whole pages missing notes are 1.29 of 3.54
points.

Units dropped as a whole are rare on the sets that have a lyrics
reference: at most 0.48 \% (CantusDB validation), 7 of 6,669 units on
whole pages. But those sets consist of manuscripts whose script the
model knows from training. The Aldersbach check set tests this failure
mode on abbreviated text (Table 6). In staff reading, P-d has a pitch
SER of 1.57 \% there, and not one wrong note: every pitch error is a
missing or an extra note, and both hang on the text. In one staff the
model leaves out the beginning of the line, five units; in another it
invents five notes for three syllables above which the manuscript has
none; in a third it writes one note too many for a wrongly divided
syllable. Seven units are dropped as a whole (0.91 \%). These figures
are subject to the correction bias described in Section 4.5.

For P-e, training examples were added in which the text band is made
illegible, for a whole line or for single words, with targets that keep
the notes and mark each unreadable syllable with \texttt{\_} (Appendix
G). The model learned them, keeping the notes under an illegible line
and placing 696 of 706 \texttt{\_} correctly, and missing notes on the
held-out manuscripts fell from 0.72 to 0.39 \%; but it was in no epoch
better than P-d on the selection sets and dropped more units on whole
pages, so P-d remains the default model. Whether such training transfers
to abbreviations, where ink is present that the model cannot interpret,
remains open, and notes dropped at abbreviated text remain an unsolved
problem.

\begin{figure}
\centering
\pandocbounded{\includegraphics[keepaspectratio,alt={Pitch errors of P-d
by kind, as shares of the reference notes. \textdagger{}Aldersbach: 55 staves of
five pages, corrected from a draft of the same model. The bar for whole
pages is computed over 25 validation and test pages, two more than the
standard measurement of Table 5.}]{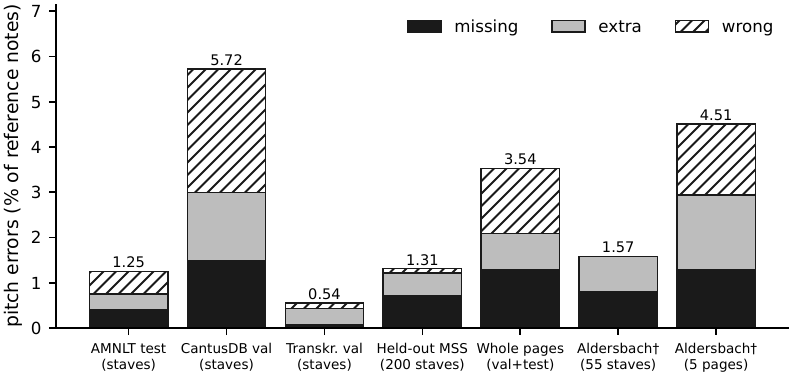}}
\caption{Pitch errors of P-d by kind, as shares of the reference notes.
\textdagger{}Aldersbach: 55 staves of five pages, corrected from a draft of the same
model. The bar for whole pages is computed over 25 validation and test
pages, two more than the standard measurement of Table 5.}
\end{figure}

\needspace{10\baselineskip}

\begin{longtable}[]{@{}
  >{\raggedright\arraybackslash}p{(\linewidth - 16\tabcolsep) * \real{0.1172}}
  >{\raggedleft\arraybackslash}p{(\linewidth - 16\tabcolsep) * \real{0.0966}}
  >{\raggedleft\arraybackslash}p{(\linewidth - 16\tabcolsep) * \real{0.1172}}
  >{\raggedleft\arraybackslash}p{(\linewidth - 16\tabcolsep) * \real{0.0966}}
  >{\raggedleft\arraybackslash}p{(\linewidth - 16\tabcolsep) * \real{0.0966}}
  >{\raggedleft\arraybackslash}p{(\linewidth - 16\tabcolsep) * \real{0.1172}}
  >{\raggedleft\arraybackslash}p{(\linewidth - 16\tabcolsep) * \real{0.1034}}
  >{\raggedleft\arraybackslash}p{(\linewidth - 16\tabcolsep) * \real{0.0966}}
  >{\raggedright\arraybackslash}p{(\linewidth - 16\tabcolsep) * \real{0.1586}}@{}}
\caption{Aldersbach check set, P-d: the 55 staves read as staff crops
and as five whole pages, against the same targets. Targets were
corrected from P-d predictions; see Section 4.5 for the resulting bias,
and Appendix A for repeated inference runs.}\tabularnewline
\toprule\noalign{}
\begin{minipage}[b]{\linewidth}\raggedright
Reading
\end{minipage} & \begin{minipage}[b]{\linewidth}\raggedleft
Pitch SER
\end{minipage} & \begin{minipage}[b]{\linewidth}\raggedleft
Missing
\end{minipage} & \begin{minipage}[b]{\linewidth}\raggedleft
Extra
\end{minipage} & \begin{minipage}[b]{\linewidth}\raggedleft
Wrong
\end{minipage} & \begin{minipage}[b]{\linewidth}\raggedleft
Dropped units
\end{minipage} & \begin{minipage}[b]{\linewidth}\raggedleft
Lyrics CER
\end{minipage} & \begin{minipage}[b]{\linewidth}\raggedleft
AMLER
\end{minipage} & \begin{minipage}[b]{\linewidth}\raggedright
Initial clef
\end{minipage} \\
\midrule\noalign{}
\endfirsthead
\toprule\noalign{}
\begin{minipage}[b]{\linewidth}\raggedright
Reading
\end{minipage} & \begin{minipage}[b]{\linewidth}\raggedleft
Pitch SER
\end{minipage} & \begin{minipage}[b]{\linewidth}\raggedleft
Missing
\end{minipage} & \begin{minipage}[b]{\linewidth}\raggedleft
Extra
\end{minipage} & \begin{minipage}[b]{\linewidth}\raggedleft
Wrong
\end{minipage} & \begin{minipage}[b]{\linewidth}\raggedleft
Dropped units
\end{minipage} & \begin{minipage}[b]{\linewidth}\raggedleft
Lyrics CER
\end{minipage} & \begin{minipage}[b]{\linewidth}\raggedleft
AMLER
\end{minipage} & \begin{minipage}[b]{\linewidth}\raggedright
Initial clef
\end{minipage} \\
\midrule\noalign{}
\endhead
\bottomrule\noalign{}
\endlastfoot
55 staff crops & 1.57 \% & 0.79 \% & 0.79 \% & 0.00 \% & 7 (0.91 \%) &
4.21 \% & 3.67 \% & F 5 of 12, C 41 of 41 \\
5 whole pages & 4.51 \% & 1.29 \% & 1.65 \% & 1.57 \% & 7 (0.91 \%) &
5.71 \% & 8.62 \% & -- \\
\end{longtable}

\subsection{Pages against staves on a difficult
manuscript}\label{pages-against-staves-on-a-difficult-manuscript}

On the same five pages, whole-page reading is clearly worse than staff
reading: 4.51 against 1.57 \% pitch SER, and unlike in staff reading
there are wrong notes (Table 6). The error is concentrated: three pages
are read at 1.4 to 3.9 \%, one at 5.2 \%, and one at 14.8 \%, on which
the page reading leaves out a staff altogether and sets a wrong clef in
two others. A dropped staff shifts every following line, so page
readings are aligned to the staves before they are scored. This
contrasts with the 20 pages of the held-out manuscripts, where page and
staff reading are level (Section 7.3), and shows that the page model is
not uniformly as good as its staff reading on material that is hard for
other reasons.

\subsection{A neighbouring staff in the
crop}\label{a-neighbouring-staff-in-the-crop}

Cropping single staves introduces an error that no training on clean
crops removes (Section 1). On 30 staves of A-Wn 15484, the first model
reads clean crops at 0.97 \% pitch SER; if the crop extends to the
middle of the next staff, at 12.67 \%; with the whole next staff, at
60.43 \%. The model also transcribes the notes of the neighbouring
staff. Image heuristics for a clean cut failed on manuscripts with tight
line spacing, and so the affected staves were discarded (Section 4.2). A
page model has no such crop.

\subsection{Text}\label{text}

The lyrics are the weaker modality on unseen manuscripts. Against the
corrected Transkribus text of the held-out manuscripts, the normalised
CER fell from 15.5 \% (S-AMNLT) to 7.9 \% (S+C10w+T), with error-free
text in 77 of 162 staves; A-KN 65 and A-Wn 15484 remained at 10 to 16
\%. Part of this is convention: the reference expands abbreviations and
often omits the letter of an initial, and a correctly read initial then
counts as an error. Part is the model: it continues a chant it knows
from training where the manuscript goes on differently, and it guesses
initials. Visual inspection suggests that the model generally follows
the image, but occasional continuations and guessed initials make its
lyrics unsuitable as ground truth; that is why the silver data take only
the notes from readings and the text from human correction.

\subsection{Shape signs}\label{shape-signs}

Shape-sign errors contribute to MER but not to pitch SER. Of the 8.26
points of MER of P-d on the Salzinnes test set, 6.82 are shape signs and
1.07 are positions. For the sign \texttt{v} alone, the prediction omits
346 reference occurrences and adds 166 occurrences absent from the
reference. On Solesmes, shape signs (3.23 points) and the neume
separator \texttt{!} (3.44) make up 6.67 of 8.92 points, positions 0.90.
Whether the model overlooks thin stems or the reference is inconsistent
was not checked at the image.

\section{Discussion: Five Hypotheses}\label{discussion-five-hypotheses}

Why does this approach read the AMNLT datasets so much more accurately
than the methods of the AMNLT study? None of the following hypotheses is
tested by a run of its own; each is given with its evidence, where
available a counter-argument, and the test that would decide. The gain
lies mostly in the content error, not in the alignment (Section 7.1), so
the hypotheses must explain why notes and text are recognised better.

\textbf{H1: OCR pre-training improves data efficiency.} A page OCR model
can already read dense pages and has only the notation to learn, whereas
the methods of the AMNLT study learn to read on each corpus. The AMNLT
study names the lack of data as the reason for the weakness of its
language-modelling methods on Solesmes \citep[Sections 8 and
9]{fuentes2026amnlt}, and the reduction in AMLER is largest there; but
we have not established how SMT and DAN were initialised. Test: compare
learning curves on growing subsets of Solesmes, with and without OCR
pre-training; H1 predicts that the pre-trained model reaches a given
error rate with fewer training examples.

\textbf{H2: The vision-language model of the AMNLT study failed in its
fine-tuning, not as a class.} PaliGemma 2 was adapted with LoRA of rank
8 in the language model, with vision block and projector frozen, on
images of 896 px \citep[Sections 7.4.2 and 7.5]{fuentes2026amnlt}; here
the vision encoder is trained in full on inputs of up to 1,540 px, with
long staves stacked in strips. A frozen encoder cannot adapt to chant
notation, and pitch reading depends strongly on resolution (Sections 5
and 7.3); scaled to 896 px with its aspect ratio kept, a staff 3,300 px
wide has about 8 px per staff step. Test: a run like S-AMNLT with the
vision encoder frozen, and one with unstacked staves.

\textbf{H3: One encoding allows joint training.} The AMNLT study trains
per corpus with separate vocabularies and names the missing common
encoding as an obstacle to pre-training across datasets; one model here
learns all corpora together. The pool of our own data cannot explain the
lead on the benchmark, which the first run already has without it; for
unseen manuscripts our results are consistent with the pool and its
encoding being what matters (Section 7.2), but pool size, coverage of
notations and encoding were never varied separately. Test: a run on
Solesmes alone.

\textbf{H4: Position letters instead of pitch names (Einsiedeln and
Salzinnes only).} The model writes the position it sees and need not
fold the clef into every note; on Salzinnes the pitch error is 1.4 \% at
a MER of 8.3 \%, the rest being mostly shape signs (Section 8.6). As a
sole explanation it fails: Solesmes is in gabc in the AMNLT study too,
and the distance is largest there. Test: a run like S-AMNLT with the
original pseudo-gabc targets for these two datasets, measured in both
encodings.

\textbf{H5: Strength in both modalities at once.} OCR pre-training may
supply the text and full training of the vision encoder the notes. The
AMNLT study describes DAN as strong in text and weak in notes and shows
a DAN prediction with perfect text and failed notes (CER 0 \%, MER 93.75
\%); our errors are low in both (P-d on Einsiedeln: MER 1.68 \%, CER
1.15 \%, against 7.23 and 3.44 \% for DAN), and a failure of the kind
shown for DAN does not occur among our test staves (Appendix F). This
checks the predicted pattern without isolating either contribution.

\section{Limitations}\label{limitations}

\emph{Single runs.} Each figure comes from one run with one seed;
differences of a few tenths of a point, in particular among P-b, P-c and
P-d, are within the variation we must assume.

\emph{The benchmark.} The comparison with the published figures is
subject to the reservations of Section 7.1; only Solesmes is directly
comparable. GregoSynth was not measured.

\emph{Confounded factors.} Architecture, pre-training on page OCR and
full training of the vision encoder are not separated in any of our
runs.

\emph{Small evaluation sets.} 40 staves per held-out manuscript and 47
initial F clefs in all; nine validation pages from three manuscripts,
two staves of which have a wrong target; the Aldersbach set has 55
staves and the biases of Section 4.5.

\emph{Text reference.} Lyrics are measured against transcriptions that
expand abbreviations (Section 4.3). Without a diplomatic transcription
that keeps them, the rate of dropped units on manuscripts rich in
abbreviations is known only from the Aldersbach set and from the
impression of the author, who judges it higher than our figures.

\emph{Shape signs.} The virga is encoded only in AMNLT (Section 3); with
the growing share of our own targets, P-d writes hardly any virga on the
held-out manuscript in square notation (47 places against 325 for P-c).
MER is not comparable across sources.

\emph{Silver data.} The two-reader filter passes errors both readers
share (Section 4.4); its measured precision is a lower bound, depressed
by errors of the ground truth, and the rate of shared errors is
estimated from one collection only.

\emph{Page reading.} On small notation the whole page falls behind its
halves, and on one difficult manuscript behind staff reading measured
with perfect boxes (Sections 7.3 and 8.3).

\emph{Scale.} About a billion parameters, which the AMNLT study itself
counts as a cost; training needs a GPU with 27 GB for half a day.

\section{Conclusion}\label{conclusion}

A fully fine-tuned page OCR model reads chant notation and lyrics
together and aligned; with the published training split alone it more
than halves the best published AMLER on Solesmes, the one directly
comparable dataset. With the architecture fixed, the largest gains on
unseen manuscripts accompanied changes to the training data and the
input layout. The consistency of the data has therefore become a central
practical constraint, alongside the recognition errors that remain: the
Gothic F clef, notes dropped with illegible or abbreviated text, and
page reading on small notation.

Controlled ablations of OCR pre-training, encoder fine-tuning, input
layout, corpus pooling and target encoding are needed to explain the
gains (Section 9), and a common encoding that also covers the shape
signs would make MER comparable across sources, the step the AMNLT study
names as the precondition for pre-training across datasets.

\section*{Acknowledgements}\label{acknowledgements}
\addcontentsline{toc}{section}{Acknowledgements}

This work was developed in the context of the collaboration between the
University of Alicante and the Austrian Centre for Digital Humanities
(ACDH), Austrian Academy of Sciences; an earlier result of that
collaboration is the tool presented in \citet{calvozaragoza2026browser}.
The AMNLT datasets, the task and its metrics are the work of the Pattern
Recognition and Artificial Intelligence Group (PRAIG) of the University
of Alicante; the training framework and the original CNN--BiLSTM--CTC
architecture of the second reader of Section 4.4 come from that group.
The common GABC encoding is defined in the GabcParser of Daniel Kurek.
Training ran on the MUSICA cluster of Austrian Scientific Computing.

The digitisation of the choir books of the Central Library of the
Franciscans in Graz was funded by the Federal Ministry for Arts,
Culture, Civil Service and Sport, ``Kulturerbe Digital'' programme, and
by the European Union.

\section*{Use of AI tools}\label{use-of-ai-tools}
\addcontentsline{toc}{section}{Use of AI tools}

The experiments were planned and decided by the author. Data preparation
scripts, training and evaluation code, the measurements and a first
draft of this text were produced with AI coding assistants working under
his direction (Anthropic Claude Fable 5.2, Opus 5.5 and OpenAI GPT 6.1
Sol); review sheets and samples were checked by him against the source
images. He takes full responsibility for the content.

\section*{Data and code availability}\label{data-and-code-availability}
\addcontentsline{toc}{section}{Data and code availability}

The code for training and inference is publicly available at
\url{https://github.com/rklugsederoeaw/lighton-choral} under the Apache
License 2.0; the repository holds code only, no ground truth. Model
weights, if released, will be licensed under CC BY-NC 4.0, because the
AMNLT datasets used in training are licensed under CC BY-NC 4.0. The
ground truth built for this article is kept private on Hugging Face,
because the licences of the material are not yet clarified.

Almost all images of the Austrian manuscripts are available at
www.cantusplanus.at; the rights are held by the Austrian Academy of
Sciences. The Klosterneuburg manuscripts (A-KN) are available through
manuscripta.at (likewise Austrian Academy of Sciences); the images of
A-Gu 17 come from the University Library of Graz.

\bibliography{refs}

\appendix

\needspace{18\baselineskip}

\section{Reproducibility}\label{reproducibility}

Software as used in run P-e (MUSICA job 1945478): container NGC PyTorch
25.04 in Apptainer with a virtual environment; Python 3.12.3, PyTorch
2.7.0a0 (NGC build nv25.04), CUDA 12.9, cuDNN 9.9.0, transformers
5.22.1, tokenizers 0.23.3, safetensors 0.8.0, huggingface\_hub 1.33.0,
NumPy 1.26.4, Pillow 11.1.0; for the metrics also rapidfuzz 3.14.6.

The published checkpoint stores the decoder weights under a different
prefix than the library version we used expects; the library reports
this as a notice only and leaves all 28 decoder layers randomly
initialised. We rename the weights before training and verify that all
532 tensors are identical bit for bit. Without that comparison the error
would have gone unnoticed, since a randomly initialised decoder also
learns, only worse. Every upload to the cluster was checked against a
manifest, byte for byte. Each job trains and then runs inference with
every checkpoint on all validation sets, the held-out manuscripts and
the page set. Inference is greedy (no sampling) in bf16 on the GPU.
Repeated inference runs occasionally differ in single characters because
of rounding differences; on the Aldersbach check set this affected 3 of
55 staves (lyrics CER 4.21 against 4.33 \%). For each model and input
layout, all Aldersbach metrics reported together were computed from the
same set of predictions.

\needspace{18\baselineskip}

\section{Held-out manuscripts by
manuscript}\label{held-out-manuscripts-by-manuscript}

\needspace{14\baselineskip}

\begin{longtable}[]{@{}
  >{\raggedright\arraybackslash}p{(\linewidth - 18\tabcolsep) * \real{0.1465}}
  >{\raggedright\arraybackslash}p{(\linewidth - 18\tabcolsep) * \real{0.1210}}
  >{\raggedleft\arraybackslash}p{(\linewidth - 18\tabcolsep) * \real{0.1083}}
  >{\raggedleft\arraybackslash}p{(\linewidth - 18\tabcolsep) * \real{0.0892}}
  >{\raggedleft\arraybackslash}p{(\linewidth - 18\tabcolsep) * \real{0.0892}}
  >{\raggedleft\arraybackslash}p{(\linewidth - 18\tabcolsep) * \real{0.0955}}
  >{\raggedleft\arraybackslash}p{(\linewidth - 18\tabcolsep) * \real{0.1210}}
  >{\raggedleft\arraybackslash}p{(\linewidth - 18\tabcolsep) * \real{0.0764}}
  >{\raggedleft\arraybackslash}p{(\linewidth - 18\tabcolsep) * \real{0.0764}}
  >{\raggedleft\arraybackslash}p{(\linewidth - 18\tabcolsep) * \real{0.0764}}@{}}
\caption{Pitch SER in \% on the 40 staves of each held-out
manuscript.}\tabularnewline
\toprule\noalign{}
\begin{minipage}[b]{\linewidth}\raggedright
Manuscript
\end{minipage} & \begin{minipage}[b]{\linewidth}\raggedright
Notation
\end{minipage} & \begin{minipage}[b]{\linewidth}\raggedleft
S-AMNLT
\end{minipage} & \begin{minipage}[b]{\linewidth}\raggedleft
S+C3
\end{minipage} & \begin{minipage}[b]{\linewidth}\raggedleft
S+C10
\end{minipage} & \begin{minipage}[b]{\linewidth}\raggedleft
S+C10w
\end{minipage} & \begin{minipage}[b]{\linewidth}\raggedleft
S+C10w+T
\end{minipage} & \begin{minipage}[b]{\linewidth}\raggedleft
P-b
\end{minipage} & \begin{minipage}[b]{\linewidth}\raggedleft
P-c
\end{minipage} & \begin{minipage}[b]{\linewidth}\raggedleft
P-d
\end{minipage} \\
\midrule\noalign{}
\endfirsthead
\toprule\noalign{}
\begin{minipage}[b]{\linewidth}\raggedright
Manuscript
\end{minipage} & \begin{minipage}[b]{\linewidth}\raggedright
Notation
\end{minipage} & \begin{minipage}[b]{\linewidth}\raggedleft
S-AMNLT
\end{minipage} & \begin{minipage}[b]{\linewidth}\raggedleft
S+C3
\end{minipage} & \begin{minipage}[b]{\linewidth}\raggedleft
S+C10
\end{minipage} & \begin{minipage}[b]{\linewidth}\raggedleft
S+C10w
\end{minipage} & \begin{minipage}[b]{\linewidth}\raggedleft
S+C10w+T
\end{minipage} & \begin{minipage}[b]{\linewidth}\raggedleft
P-b
\end{minipage} & \begin{minipage}[b]{\linewidth}\raggedleft
P-c
\end{minipage} & \begin{minipage}[b]{\linewidth}\raggedleft
P-d
\end{minipage} \\
\midrule\noalign{}
\endhead
\bottomrule\noalign{}
\endlastfoot
A-Wn 15484 & square & 0.72 & 1.43 & 1.29 & 1.14 & 0.86 & 1.14 & 1.14 &
2.43 \\
A-Geras 6 & German Gothic & 18.75 & 4.03 & 1.73 & 1.81 & 1.89 & 1.81 &
0.90 & 0.49 \\
A-KN 65 & Bohemian Gothic & 10.79 & 8.25 & 7.36 & 6.22 & 3.68 & 3.43 &
2.66 & 2.28 \\
A-Gu 17 & German Gothic & 46.36 & 10.01 & 8.25 & 8.25 & 2.24 & 1.36 &
0.96 & 1.28 \\
A-VOR 254 & German Gothic & 34.41 & 23.41 & 10.14 & 4.96 & 1.83 & 1.19 &
1.08 & 0.76 \\
all & & 24.92 & 9.55 & 5.84 & 4.67 & 2.11 & 1.74 & 1.27 & 1.31 \\
\end{longtable}

The rise for A-Wn 15484 with P-d (2.43 \%) comes mostly from the one
staff discussed in Section 8.2, whose eight missing notes are 1.1 points
of that manuscript's 699 notes.

\needspace{18\baselineskip}

\section{All epochs of the page
models}\label{all-epochs-of-the-page-models}

\needspace{12\baselineskip}

\begin{longtable}[]{@{}
  >{\raggedright\arraybackslash}p{(\linewidth - 10\tabcolsep) * \real{0.1176}}
  >{\raggedleft\arraybackslash}p{(\linewidth - 10\tabcolsep) * \real{0.1176}}
  >{\raggedleft\arraybackslash}p{(\linewidth - 10\tabcolsep) * \real{0.1933}}
  >{\raggedleft\arraybackslash}p{(\linewidth - 10\tabcolsep) * \real{0.1933}}
  >{\raggedleft\arraybackslash}p{(\linewidth - 10\tabcolsep) * \real{0.1933}}
  >{\raggedleft\arraybackslash}p{(\linewidth - 10\tabcolsep) * \real{0.1849}}@{}}
\caption{Pitch SER in \% and validation loss by epoch.}\tabularnewline
\toprule\noalign{}
\begin{minipage}[b]{\linewidth}\raggedright
Model
\end{minipage} & \begin{minipage}[b]{\linewidth}\raggedleft
Epoch
\end{minipage} & \begin{minipage}[b]{\linewidth}\raggedleft
Validation loss
\end{minipage} & \begin{minipage}[b]{\linewidth}\raggedleft
AMNLT validation
\end{minipage} & \begin{minipage}[b]{\linewidth}\raggedleft
9 validation pages
\end{minipage} & \begin{minipage}[b]{\linewidth}\raggedleft
Held-out, 200 staves
\end{minipage} \\
\midrule\noalign{}
\endfirsthead
\toprule\noalign{}
\begin{minipage}[b]{\linewidth}\raggedright
Model
\end{minipage} & \begin{minipage}[b]{\linewidth}\raggedleft
Epoch
\end{minipage} & \begin{minipage}[b]{\linewidth}\raggedleft
Validation loss
\end{minipage} & \begin{minipage}[b]{\linewidth}\raggedleft
AMNLT validation
\end{minipage} & \begin{minipage}[b]{\linewidth}\raggedleft
9 validation pages
\end{minipage} & \begin{minipage}[b]{\linewidth}\raggedleft
Held-out, 200 staves
\end{minipage} \\
\midrule\noalign{}
\endhead
\bottomrule\noalign{}
\endlastfoot
P-b & 1 & 0.1228 & 3.12 & 23.4 & 9.41 \\
P-b & 2 & 0.0984 & 1.96 & 11.2 & 1.50 \\
P-b & 3 & 0.0822 & 1.65 & 5.4 & 1.46 \\
P-b & 4 & 0.1014 & 1.71 & 2.9 & 1.74 \\
P-c & 1 & 0.1217 & 2.35 & 16.51 & -- \\
P-c & 2 & 0.0888 & 1.92 & 8.36 & -- \\
P-c & 3 & 0.0929 & 1.79 & 4.67 & -- \\
P-c & 4 & 0.1119 & 1.77 & 3.98 & 1.27 \\
P-d & 1 & 0.1086 & 2.69 & 11.42 & 3.09 \\
P-d & 2 & 0.0918 & 1.83 & 9.48 & 3.09 \\
P-d & 3 & 0.0916 & 1.86 & 3.86 & 1.52 \\
P-d & 4 & 0.1128 & 1.74 & 3.34 & 1.31 \\
P-e & 1 & 0.1107 & 2.29 & 9.24 & 1.11 \\
P-e & 2 & 0.0886 & 2.68 & 8.41 & 3.05 \\
P-e & 3 & 0.0911 & 1.75 & 3.58 & 1.93 \\
P-e & 4 & 0.1107 & 1.86 & 3.60 & 1.62 \\
\end{longtable}

\needspace{18\baselineskip}

\section{The second reader and the ground-truth
check}\label{the-second-reader-and-the-ground-truth-check}

The second reader of Section 4.4 consists of four convolutional blocks
and a two-layer bidirectional LSTM (BiLSTM) with a CTC head and, in
parallel, a six-layer autoregressive Transformer decoder with
cross-attention, about 14.3 million parameters
\citep{klugseder2026hybrid}. Its training framework and the original
CNN--BiLSTM--CTC architecture come from the PRAIG group; the hybrid
CTC/attention extension was added in the project of the present author.

Of the 103 staves of the 14 Graz volumes on which the two readers agree
strictly and the human ground truth differs, the author checked 78
against the source images: in 68 the readers are right, in 8 the ground
truth is right, and 2 received a corrected line of his own. For the
remaining 25, all of them pure substitutions of notes at the same place,
he decided to follow the readers by rule. Of the 56 checked staves that
differ by pure substitution, the readers are right in 55 and the ground
truth in none; of the 18 staves in which the number of notes differs,
the ground truth is right in 6; of the 4 that differ in the clef only,
in 2.

\needspace{18\baselineskip}

\section{Runs in full}\label{runs-in-full}

\needspace{15\baselineskip}

\begin{longtable}[]{@{}
  >{\raggedright\arraybackslash}p{(\linewidth - 14\tabcolsep) * \real{0.1038}}
  >{\raggedright\arraybackslash}p{(\linewidth - 14\tabcolsep) * \real{0.1257}}
  >{\raggedright\arraybackslash}p{(\linewidth - 14\tabcolsep) * \real{0.3388}}
  >{\raggedleft\arraybackslash}p{(\linewidth - 14\tabcolsep) * \real{0.0820}}
  >{\raggedleft\arraybackslash}p{(\linewidth - 14\tabcolsep) * \real{0.0765}}
  >{\raggedleft\arraybackslash}p{(\linewidth - 14\tabcolsep) * \real{0.1038}}
  >{\raggedleft\arraybackslash}p{(\linewidth - 14\tabcolsep) * \real{0.0929}}
  >{\raggedleft\arraybackslash}p{(\linewidth - 14\tabcolsep) * \real{0.0765}}@{}}
\caption{Runs in full. Pitch SER in \%, held-out: 200 staves. Initial
clef on the held-out manuscripts. Text CER: normalised (case, u/v, i/j,
letters only) against the corrected Transkribus text of the held-out
manuscripts; for the page models that figure was not computed. Internal
names are those of the repository.}\tabularnewline
\toprule\noalign{}
\begin{minipage}[b]{\linewidth}\raggedright
Name
\end{minipage} & \begin{minipage}[b]{\linewidth}\raggedright
Internal
\end{minipage} & \begin{minipage}[b]{\linewidth}\raggedright
Training data (examples per epoch)
\end{minipage} & \begin{minipage}[b]{\linewidth}\raggedleft
Chosen epoch
\end{minipage} & \begin{minipage}[b]{\linewidth}\raggedleft
AMNLT val
\end{minipage} & \begin{minipage}[b]{\linewidth}\raggedleft
Held-out
\end{minipage} & \begin{minipage}[b]{\linewidth}\raggedleft
Initial clef
\end{minipage} & \begin{minipage}[b]{\linewidth}\raggedleft
Text CER
\end{minipage} \\
\midrule\noalign{}
\endfirsthead
\toprule\noalign{}
\begin{minipage}[b]{\linewidth}\raggedright
Name
\end{minipage} & \begin{minipage}[b]{\linewidth}\raggedright
Internal
\end{minipage} & \begin{minipage}[b]{\linewidth}\raggedright
Training data (examples per epoch)
\end{minipage} & \begin{minipage}[b]{\linewidth}\raggedleft
Chosen epoch
\end{minipage} & \begin{minipage}[b]{\linewidth}\raggedleft
AMNLT val
\end{minipage} & \begin{minipage}[b]{\linewidth}\raggedleft
Held-out
\end{minipage} & \begin{minipage}[b]{\linewidth}\raggedleft
Initial clef
\end{minipage} & \begin{minipage}[b]{\linewidth}\raggedleft
Text CER
\end{minipage} \\
\midrule\noalign{}
\endhead
\bottomrule\noalign{}
\endlastfoot
S-AMNLT & \texttt{s4} & AMNLT (3,923) & 3 & 1.77 & 24.92 & 80.5 \% &
15.51 \\
S+C3 & \texttt{s4k} & + 2,160 CantusDB staves, three manuscripts (6,083)
& 4 & 1.79 & 9.55 & 89.5 \% & 12.17 \\
S+C10 & \texttt{s4g} & 8,781 CantusDB staves, ten manuscripts, in place
of the 2,160 (12,704) & 3 & 1.86 & 5.84 & 96.5 \% & 9.57 \\
S+C10w & \texttt{s4w} & as before, AMNLT with word boundaries (12,704) &
4 & 1.68 & 4.67 & 88.5 \% & 8.34 \\
S+C10w+T & \texttt{s4t} & + 569 staves, four Gothic manuscripts (13,273)
& 4 & 1.80 & 2.11 & 92.5 \% & 7.90 \\
P-ft & \texttt{seiten\_a} & previous model continued, + 2,390
multi-staff images (15,663) & 1 & 1.89 & 2.28 & 92.0 \% & -- \\
P-b & \texttt{seiten\_b} & + 4,551 multi-staff images (17,824) & 4 &
1.71 & 1.74 & 87.0 \% & -- \\
P-c & \texttt{seiten\_c} & cleaned Transkribus targets, Graz corpus,
F-clef staves twice (22,737) & 4 & 1.77 & 1.27 & 91.5 \% & -- \\
P-d & \texttt{seiten\_d} & + gold of 14 Graz volumes, silver, pages
three times (29,627) & 4 & 1.74 & 1.31 & 91.0 \% & -- \\
P-e & \texttt{seiten\_e} & + text-dropout examples (1,598), rebuilt
Transkribus corpora (31,393) & 3 & 1.75 & 1.93 & 91.5 \% & -- \\
4B & \texttt{b4\_full} & LightOnOCR-3-4B, data of S+C10w+T (13,273) & 4
& 1.80 & 2.38 & 91.5 \% & 8.14 \\
\end{longtable}

\needspace{18\baselineskip}

\section{The AMNLT test split in
full}\label{the-amnlt-test-split-in-full}

\needspace{15\baselineskip}

\begin{longtable}[]{@{}
  >{\raggedright\arraybackslash}p{(\linewidth - 18\tabcolsep) * \real{0.1379}}
  >{\raggedright\arraybackslash}p{(\linewidth - 18\tabcolsep) * \real{0.2299}}
  >{\raggedleft\arraybackslash}p{(\linewidth - 18\tabcolsep) * \real{0.0805}}
  >{\raggedleft\arraybackslash}p{(\linewidth - 18\tabcolsep) * \real{0.0805}}
  >{\raggedleft\arraybackslash}p{(\linewidth - 18\tabcolsep) * \real{0.0805}}
  >{\raggedleft\arraybackslash}p{(\linewidth - 18\tabcolsep) * \real{0.0805}}
  >{\raggedleft\arraybackslash}p{(\linewidth - 18\tabcolsep) * \real{0.0805}}
  >{\raggedleft\arraybackslash}p{(\linewidth - 18\tabcolsep) * \real{0.0805}}
  >{\raggedleft\arraybackslash}p{(\linewidth - 18\tabcolsep) * \real{0.0805}}
  >{\raggedleft\arraybackslash}p{(\linewidth - 18\tabcolsep) * \real{0.0690}}@{}}
\caption{AMNLT test split (562 staves) with all measures, in \% except
AlER. Published values from Tables 2 and 3 of \citet{fuentes2026amnlt};
their bWER is derived as described in Section 7.1. *Einsiedeln and
Salzinnes: different encoding and denominator, see reservation 2 in
Section 7.1.}\tabularnewline
\toprule\noalign{}
\begin{minipage}[b]{\linewidth}\raggedright
Dataset
\end{minipage} & \begin{minipage}[b]{\linewidth}\raggedright
Model
\end{minipage} & \begin{minipage}[b]{\linewidth}\raggedleft
MER
\end{minipage} & \begin{minipage}[b]{\linewidth}\raggedleft
Pitch SER
\end{minipage} & \begin{minipage}[b]{\linewidth}\raggedleft
CER
\end{minipage} & \begin{minipage}[b]{\linewidth}\raggedleft
SylER
\end{minipage} & \begin{minipage}[b]{\linewidth}\raggedleft
AMLER
\end{minipage} & \begin{minipage}[b]{\linewidth}\raggedleft
bWER
\end{minipage} & \begin{minipage}[b]{\linewidth}\raggedleft
AMLER \ensuremath{-} bWER
\end{minipage} & \begin{minipage}[b]{\linewidth}\raggedleft
AlER
\end{minipage} \\
\midrule\noalign{}
\endfirsthead
\toprule\noalign{}
\begin{minipage}[b]{\linewidth}\raggedright
Dataset
\end{minipage} & \begin{minipage}[b]{\linewidth}\raggedright
Model
\end{minipage} & \begin{minipage}[b]{\linewidth}\raggedleft
MER
\end{minipage} & \begin{minipage}[b]{\linewidth}\raggedleft
Pitch SER
\end{minipage} & \begin{minipage}[b]{\linewidth}\raggedleft
CER
\end{minipage} & \begin{minipage}[b]{\linewidth}\raggedleft
SylER
\end{minipage} & \begin{minipage}[b]{\linewidth}\raggedleft
AMLER
\end{minipage} & \begin{minipage}[b]{\linewidth}\raggedleft
bWER
\end{minipage} & \begin{minipage}[b]{\linewidth}\raggedleft
AMLER \ensuremath{-} bWER
\end{minipage} & \begin{minipage}[b]{\linewidth}\raggedleft
AlER
\end{minipage} \\
\midrule\noalign{}
\endhead
\bottomrule\noalign{}
\endlastfoot
Solesmes & published, best AMLER (unfolding CRNN) & 18.20 & -- & 19.11 &
38.88 & 20.98 & 13.01 & 7.97 & 0.38 \\
Solesmes & published, DAN & 21.87 & -- & 17.94 & 30.19 & 23.17 & 14.13 &
9.04 & 0.39 \\
Solesmes & S-AMNLT & 9.63 & 1.87 & 5.54 & 13.98 & 9.92 & 6.64 & 3.28 &
0.33 \\
Solesmes & S+C10w+T & 9.58 & 1.50 & 5.26 & 11.50 & 9.35 & 5.91 & 3.44 &
0.36 \\
Solesmes & P-d & 8.92 & 1.23 & 5.10 & 10.91 & 8.80 & 5.61 & 3.19 &
0.36 \\
Einsiedeln* & published, best AMLER (DAN) & 7.23 & -- & 3.44 & 7.19 &
5.14 & 4.27 & 0.87 & 0.17 \\
Einsiedeln* & S-AMNLT & 1.93 & 1.23 & 2.36 & 4.78 & 2.25 & 1.56 & 0.69 &
0.30 \\
Einsiedeln* & S+C10w+T & 1.67 & 1.00 & 1.60 & 3.42 & 1.84 & 1.19 & 0.65
& 0.34 \\
Einsiedeln* & P-d & 1.68 & 1.10 & 1.15 & 2.84 & 1.87 & 1.10 & 0.77 &
0.41 \\
Salzinnes* & published, best AMLER (SMT) & 13.73 & -- & 3.69 & 7.80 &
7.32 & 6.30 & 1.02 & 0.14 \\
Salzinnes* & published, DAN & 29.46 & -- & 12.34 & 21.08 & 18.73 & 14.05
& 4.68 & 0.25 \\
Salzinnes* & S-AMNLT & 8.12 & 1.63 & 1.56 & 4.06 & 5.31 & 4.47 & 0.84 &
0.15 \\
Salzinnes* & S+C10w+T & 8.20 & 1.32 & 1.19 & 3.47 & 4.54 & 3.86 & 0.68 &
0.14 \\
Salzinnes* & P-d & 8.26 & 1.41 & 0.92 & 2.74 & 4.36 & 3.53 & 0.83 &
0.18 \\
\end{longtable}

Among the 562 test staves, staves with a CER of at most 2 \% and a MER
of at least 50 \% number 2 for S+C3, 3 for S+C10w+T and 1 for P-b; the
opposite case never occurs. None of these is a failure to read pitch: in
two the pitch is right and only the sign \texttt{v} differs, and in the
third the reference has a clef that displaces all notes by four steps
while the model writes a consistent line under \texttt{c4}. A failure of
the kind the AMNLT study shows for DAN, perfect text with failed notes,
does not occur.

\needspace{18\baselineskip}

\section{The text-dropout run P-e}\label{the-text-dropout-run-p-e}

The training examples of P-e make the text band illegible in the ways
manuscripts are (smudged, faded, stained, abraded), for a whole line
(kind A) or for single words (kind B); the targets keep the notes and
write each unreadable syllable as \texttt{\_}. They make up 5.1 \% of
the examples per epoch (1,598, each example twice), alongside the
rebuilt corpora of Section 4.3 and further gold staves (31,393 examples
per epoch in all). Table 11 compares the chosen epoch 3 with P-d.

\needspace{12\baselineskip}

\begin{longtable}[]{@{}
  >{\raggedright\arraybackslash}p{(\linewidth - 4\tabcolsep) * \real{0.7000}}
  >{\raggedleft\arraybackslash}p{(\linewidth - 4\tabcolsep) * \real{0.1500}}
  >{\raggedleft\arraybackslash}p{(\linewidth - 4\tabcolsep) * \real{0.1500}}@{}}
\caption{P-e (epoch 3) against P-d.~Pitch SER in \% unless
stated.}\tabularnewline
\toprule\noalign{}
\begin{minipage}[b]{\linewidth}\raggedright
Measure
\end{minipage} & \begin{minipage}[b]{\linewidth}\raggedleft
P-d
\end{minipage} & \begin{minipage}[b]{\linewidth}\raggedleft
P-e
\end{minipage} \\
\midrule\noalign{}
\endfirsthead
\toprule\noalign{}
\begin{minipage}[b]{\linewidth}\raggedright
Measure
\end{minipage} & \begin{minipage}[b]{\linewidth}\raggedleft
P-d
\end{minipage} & \begin{minipage}[b]{\linewidth}\raggedleft
P-e
\end{minipage} \\
\midrule\noalign{}
\endhead
\bottomrule\noalign{}
\endlastfoot
Kind-A validation lines, pitch SER & -- & 0.23 \\
Kind-A validation lines, \texttt{\_} at the right place & -- & 696 of
706 \\
Held-out manuscripts, missing notes & 0.72 & 0.39 \\
AMNLT validation & 1.74 & 1.75 \\
Validation pages & 3.34 & 3.58 \\
Whole pages, dropped units & 0.11 & 0.85 \\
\texttt{\_} written where the text is legible (epochs 3 and 4) & -- & 12
to 19 \\
Aldersbach, staff crops & 1.57 & 2.22 \\
Aldersbach, whole pages & 4.51 & 3.65 \\
AMNLT test, AMLER Solesmes & 8.80 & 9.09 \\
AMNLT test, AMLER Einsiedeln & 1.87 & 1.83 \\
AMNLT test, AMLER Salzinnes & 4.36 & 4.25 \\
\end{longtable}

The wrong \texttt{\_} stand mostly at decorated initials at the start of
a staff. On the selection sets P-e is in no epoch better than P-d.

\end{document}